\documentclass[lettersize,journal]{IEEEtran}
\usepackage{amsmath,amsfonts}
\usepackage{algorithmic}
\usepackage{algorithm}
\usepackage{array}
\usepackage[caption=false,font=normalsize,labelfont=sf,textfont=sf]{subfig}
\usepackage{textcomp}
\usepackage{stfloats}
\usepackage{hyperref}
\usepackage{url}
\usepackage{verbatim}
\usepackage{graphicx}
\usepackage{cite}
\usepackage{booktabs}
\begin{document}

\title{Modular RAG: Transforming RAG Systems into LEGO-like Reconfigurable Frameworks}

\author{Yunfan Gao, Yun Xiong, Meng Wang, Haofen Wang

\thanks{
Yunfan Gao is with Shanghai Research Institute for Intelligent Autonomous Systems, Tongji University, Shanghai, 201210, China.

Yun Xiong is with Shanghai Key Laboratory of Data Science, School of Computer Science, Fudan University,  Shanghai, 200438, China. 

Meng Wang and Haofen Wang are with College of Design and Innovation, Tongji University, Shanghai, 20092, China. 
(Corresponding author: Haofen Wang. E-mail: carter.whfcarter@gmail.com)
}
}

\maketitle

\begin{abstract}
Retrieval-augmented Generation (RAG) has markedly enhanced the capabilities of Large Language Models (LLMs) in tackling knowledge-intensive tasks. The increasing demands of application scenarios have driven the evolution of RAG, leading to the integration of advanced retrievers, LLMs and other complementary technologies,  which in turn has amplified the intricacy of RAG systems. However, the rapid advancements are outpacing the foundational RAG paradigm, with many methods struggling to be unified under the process of ``retrieve-then-generate". In this context, this paper examines the limitations of the existing RAG paradigm and introduces the modular RAG framework. By decomposing complex RAG systems into independent modules and specialized operators, it facilitates a highly reconfigurable framework. Modular RAG transcends the traditional linear architecture, embracing a more advanced design that integrates routing, scheduling, and fusion  mechanisms. Drawing on extensive research, this paper further identifies prevalent RAG patterns—linear, conditional, branching, and looping—and offers a comprehensive analysis of their respective implementation nuances. Modular RAG presents innovative opportunities for the conceptualization and deployment of RAG systems. Finally, the paper explores the potential emergence of new operators and paradigms, establishing a solid theoretical foundation and a practical roadmap for the continued evolution and practical deployment of RAG technologies.
\end{abstract}


\section{Introduction}
\IEEEPARstart{L}{arge} Language Models (LLMs) have demonstrated remarkable capabilities, yet they still face numerous challenges, such as hallucination and the lag in information updates~\cite{hallucination}. Retrieval-augmented Generation (RAG), by accessing external knowledge bases, provides LLMs with important contextual information, significantly enhancing their performance on  knowledge-intensive tasks~\cite{rag_survey}. Currently, RAG, as an enhancement method, has been widely applied in various  practical application scenarios, including knowledge question answering, recommendation systems, customer service, and personal assistants.~\cite{linkedin,NoteLLM,CT-RAG,chatRec}

\begin{figure}[htbp]
    \centering
    \includegraphics[width=\linewidth]{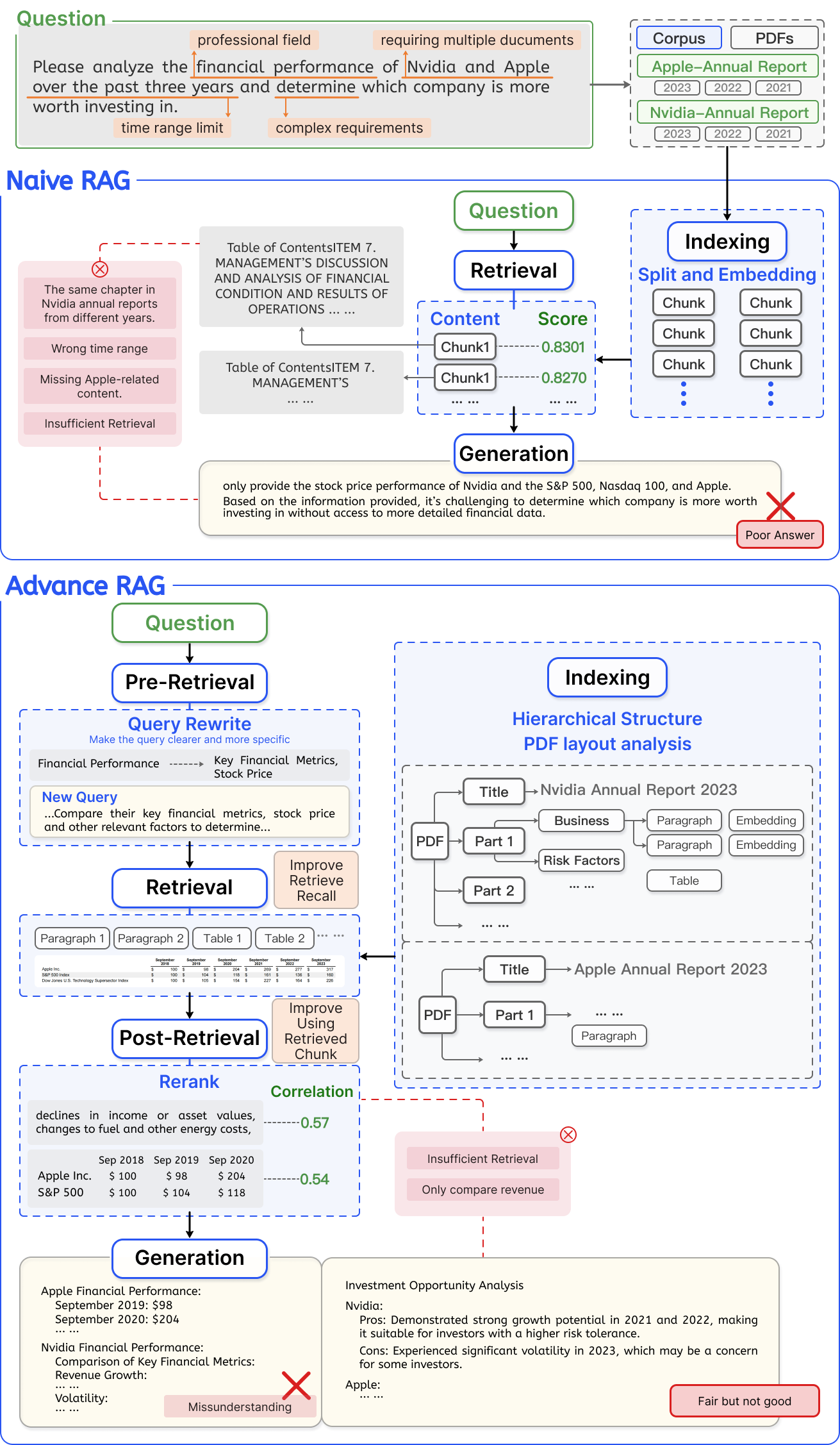}
    \caption{Cases of Naive RAG and Advanced RAG.When faced with complex questions, both encounter limitations and struggle to provide satisfactory answers. Despite the fact that Advanced RAG improves retrieval accuracy through hierarchical indexing, pre-retrieval, and post-retrieval processes, these relevant documents have not been used correctly.}
    \label{fig:base_case}
\end{figure}

During the nascent stages of RAG , its core framework is constituted by  indexing, retrieval, and generation, a paradigm referred to as Naive RAG~\cite{LlamaIndexTalk}. However, as the complexity of tasks and the demands of applications have escalated, the limitations of Naive RAG have become increasingly apparent.  As depicted in Figure~\ref{fig:base_case}, it predominantly hinges on the straightforward similarity of  chunks, result  in poor performance when confronted with complex queries and chunks with substantial variability. The primary challenges of Naive RAG include:
\emph{1) Shallow Understanding of Queries.} The semantic similarity between a query and document chunk is not always highly consistent. Relying solely on similarity calculations for retrieval lacks an in-depth exploration of the relationship between the query and the document~\cite{embedding_survey}.
\emph{2) Retrieval Redundancy and Noise.} Feeding all retrieved chunks directly into LLMs is not always beneficial. Research indicates that an excess of redundant and noisy information may interfere with the LLM's identification of key information, thereby increasing the risk of generating erroneous and hallucinated responses.~\cite{NoiseRAG}

To overcome the aforementioned limitations, Advanced RAG paradigm focuses on optimizing the retrieval phase, aiming to enhance retrieval efficiency and strengthen the utilization of retrieved chunks. As shown in Figure~\ref{fig:base_case} ,typical  strategies involve pre-retrieval processing and post-retrieval processing. For instance, query rewriting is used to make the queries more clear and specific, thereby increasing the accuracy of retrieval~\cite{BEQUE}, and the reranking of retrieval results is employed to enhance the LLM’s ability to identify and utilize key information~\cite{rerank}.

Despite the improvements in the practicality of Advanced RAG, there remains a gap between its capabilities and real-world application requirements. On one hand, as RAG technology advances, user expectations rise, demands continue to evolve, and application settings become more complex. For instance, the integration of heterogeneous data  and the new demands for system transparency, control, and maintainability. On the other hand, the growth in application demands has further propelled the evolution of RAG technology.

\begin{figure}[htbp]
    \centering
    \includegraphics[width=\linewidth]{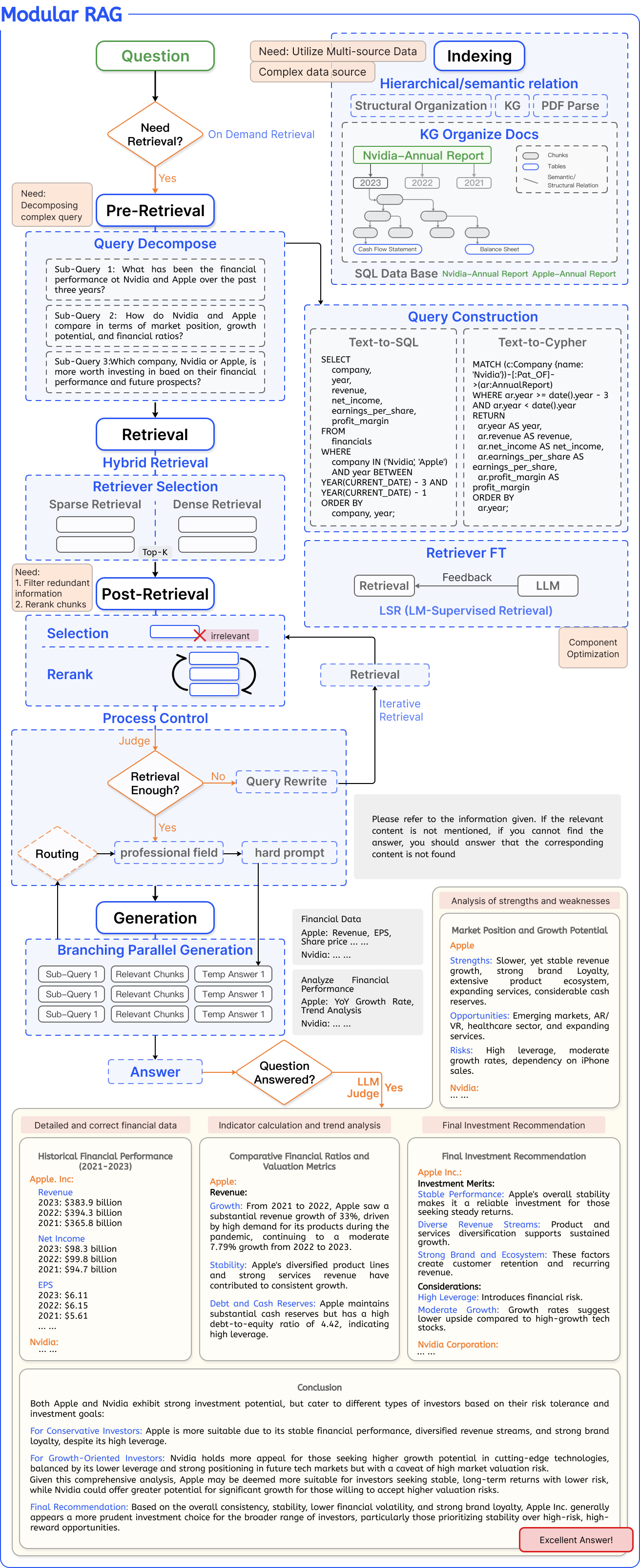}
    \caption{Case of current Modular RAG.The system integrates diverse data and more functional components. The process is no longer confined to  linear but is controlled by multiple control components for retrieval and generation, making the entire system more flexible and complex.}
    \label{fig:modular_case}
\end{figure}

As shown in Figure~\ref{fig:modular_case}, to achieve more accurate and efficient task execution, modern RAG systems are progressively integrating more sophisticated function, such as organizing more refined index base in the form of knowledge graphs, integrating structured data through query construction methods, and employing fine-tuning techniques to enable encoders to better adapt to domain-specific documents.

In terms of process design, the current RAG system has surpassed the traditional linear retrieval-generation paradigm. Researchers use iterative retrieval~\cite{ITRG} to obtain richer context, recursive retrieval~\cite{TOC} to handle complex queries, and adaptive retrieval~\cite{Flare} to provide overall autonomy and flexibility. This flexibility in the process significantly enhances the expressive power and adaptability of RAG systems, enabling them to better adapt to various application scenarios. However, this also makes the orchestration and scheduling of workflows more complex, posing greater challenges to system design. Specifically, RAG currently faces the following new challenges:

\textbf{Complex data sources integration.} RAG  are no longer confined to a single type of unstructured text data source but have expanded to include various data types, such as semi-structured data like tables and structured data like knowledge graphs~\cite{graphRAG}. Access to heterogeneous data from multiple sources can provide the system with a richer knowledge background,  and more reliable knowledge verification capabilities~\cite{Databricks-RAG}.

 \textbf{New demands for system interpretability, controllability, and maintainability.} With the increasing complexity of systems, system maintenance and debugging have become more challenging. Additionally, when issues arise, it is essential to quickly pinpoint the specific components that require optimization.
 
\textbf{Component selection and optimization.} More neural networks are involved in the RAG system, necessitating the selection of appropriate components to meet the needs of specific tasks and resource configurations. Moreover, additional components enhance the effectiveness of RAG but also bring new collaborative work requirements~\cite{searchRAG}. Ensuring that these models perform as intended and work efficiently together to enhance the overall system performance is crucial.

\textbf{Workflow orchestration and scheduling.} Components may need to be executed in a specific order, processed in parallel under certain conditions, or even judged by the LLM based on different outputs. Reasonable planning of the workflow is essential for improving system efficiency and achieving the desired outcomes~\cite{planrag}.

To address the design, management, and maintenance challenges posed by the increasing complexity of RAG systems, and to meet the ever-growing and diverse demands and expectations, this paper proposes Modular RAG architecture. In modern computing systems, modularization is becoming a trend. It can enhance the system's scalability and maintainability and achieve efficient task execution through process control.

\begin{figure*}
	\centering
		\includegraphics[width= \textwidth]{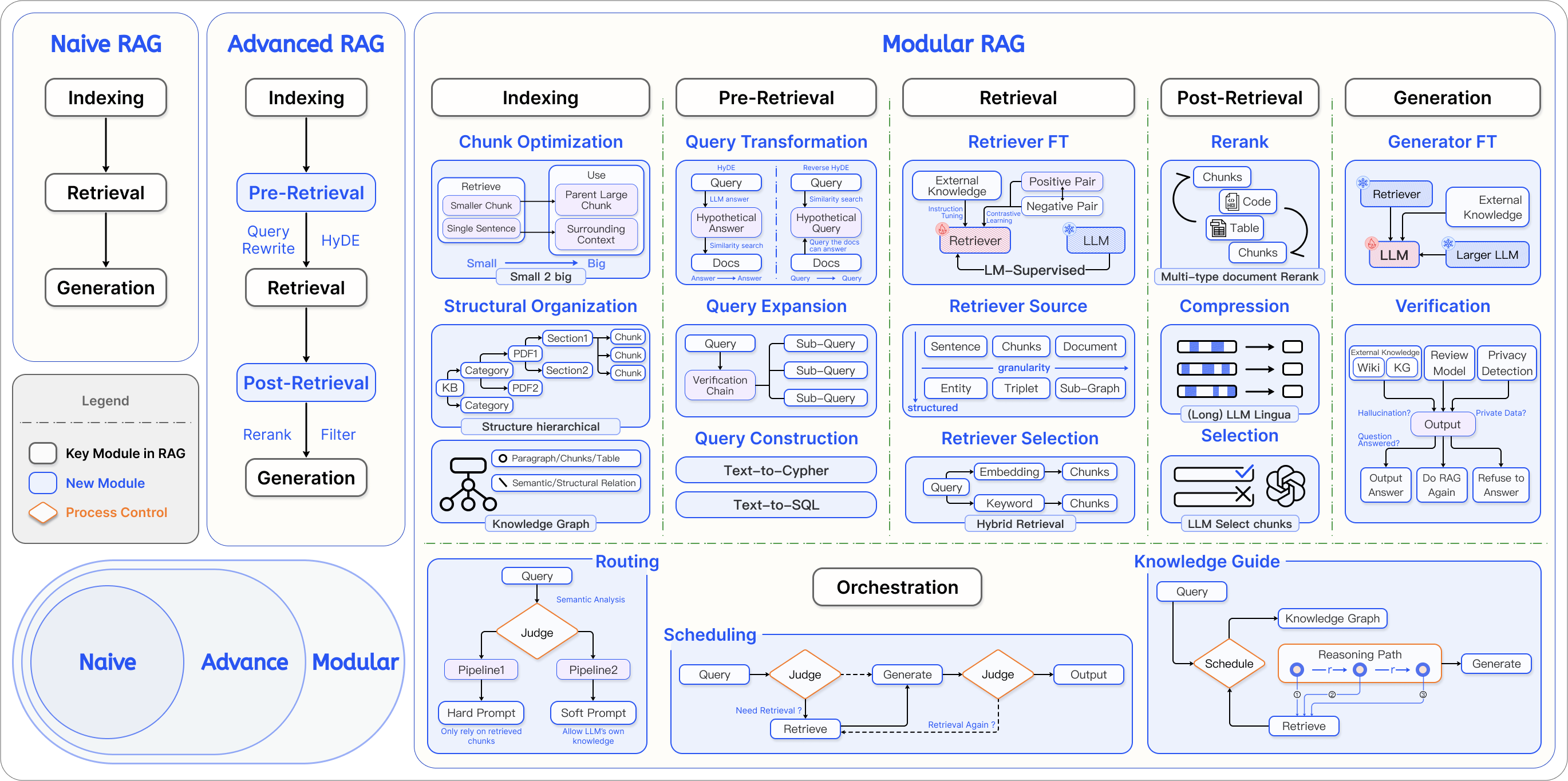}
	\caption{Comparison between three RAG paradigms. Modular RAG has evolved from previous paradigms and aligns with the current practical needs of RAG systems.}
	\label{fig:Compare}
\end{figure*}

The Modular RAG system consists of multiple independent yet tightly coordinated modules, each responsible for handling specific functions or tasks. This architecture is divided into three levels: the top level focuses on the critical stages of RAG, where each stage is treated as an independent module. This level not only inherits the main processes from the Advanced RAG paradigm but also introduces an orchestration module to control the coordination of RAG processes. The middle level is composed of sub-modules within each module, further refining and optimizing the functions. The bottom level consists of basic units of operation—operators. Within the Modular RAG framework, RAG systems can be represented in the form of computational graphs, where nodes represent specific operators. The comparison of the three paradigms is shown in the Figure~\ref{fig:Compare}. Modular RAG evolves based on the previous development of RAG. The relationships among these three paradigms are ones of inheritance and development. Advanced RAG is a special case of Modular RAG, while Naive RAG is a special case of Advanced RAG.

The advantages of Modular RAG are significant, as it enhances the flexibility and scalability of RAG systems.  Users can flexibly combine different modules and operators according to the requirements of data sources and task scenarios. In summary, the contributions of this paper are as follows:
\begin{itemize}
    \item This paper proposes a new paradigm called modular RAG, which employs a three-tier architectural design comprising modules, sub-modules, and operators to define the RAG system in a unified and structured manner. This design not only enhances the system's flexibility and scalability but also, through the independent design of operators, strengthens the system's maintainability and comprehensibility.
    \item Under the framework of Modular RAG, the orchestration of modules and operators forms the RAG Flow, which can flexibly express current RAG methods. This paper has further summarized six typical flow patterns and specific methods have been analyzed to reveal the universality of modular RAG in practical scenarios.
    \item The Modular RAG framework offers exceptional flexibility and extensibility. This paper delves into the new opportunities brought by Modular RAG and provides a thorough discussion on the adaptation and expansion of new methods in different application scenarios, offering guidance for future research directions and practical exploration.
\end{itemize}

\section{Related Work}
The development of RAG technology can be summarized in three stages. Initially, retrieval-augmented techniques were introduced to improve the performance of pre-trained language models on knowledge-intensive tasks~\cite{REALM,RAG}. In specific implementations, Retro~\cite{Retro} optimized pre-trained autoregressive models through retrieval augmentation, while Atlas~\cite{Atlas} utilized a retrieval-augmented few-shot fine-tuning method, enabling language models to adapt to diverse tasks. IRCOT~\cite{IRCoT} further enriched the reasoning process during the inference phase by combining chain-of-thought and multi-step retrieval processes. Entering the second stage, as the language processing capabilities of LLMs significantly improved, retrieval-augmented techniques began to serve as a means of supplementing additional knowledge and providing references, aiming to reduce the hallucination. For instance, RRR~\cite{RRR} improved the rewriting phase, and LLMlingua~\cite{LLMLingua} removed redundant tokens in retrieved document chunks. With the continuous progress of RAG technology, research has become more refined and focused, while also achieving innovative integration with other technologies such as graph neural networks~\cite{RoG} and fine-tuning techniques~\cite{RA-DIT}. The overall  pipeline has also become more flexible, such as using LLMs to proactively determine the timing of retrieval and generation~\cite{Flare,self-rag}.

The development of RAG technology has been accelerated by LLM technology and practical application needs. Researchers are examining and organizing the RAG framework and development pathways from different perspectives. Building upon the enhanced stages of RAG, Gao et al.,~\cite{rag_survey} subdivided RAG into enhancement during pre-training, inference, and fine-tuning stages. Based on the main processes of RAG, relevant works on RAG were organized from the perspectives of retrieval, generation, and augmentation methods. Huang et al.,~\cite{RAG_text} categorize RAG methods into four main classes: pre-retrieval, retrieval, post-retrieval, generation, and provide a detailed discussion of the methods and techniques within each class. Hu et al.,~\cite{rag_rau} discuss Retrieval-Augmented Language Models (RALMs) form three key components, including retrievers, language models, augmentations, and how their interactions lead to different model structures and applications. They emphasize the importance of considering robustness, accuracy, and relevance when evaluating RALMs and propose several evaluation methods. Ding et al.,~\cite{ragMeet} provide a comprehensive review  from the perspectives of architecture, training strategies, and applications. They specifically discuss four training methods of RALMs: training-free methods, independent training methods, sequence training methods, and joint training methods, and compare their advantages and disadvantages. Zhao et al.,~\cite{RAG4AIGC}analyze the applications of RAG technology in various fields such as text generation, code generation, image generation, and video generation from the perspective of augmented intelligence with generative capabilities.

The current collation of RAG systems primarily focuses on methods with a fixed process, mainly concerned with optimizing the retrieval and generation stages. However, it has not turned its attention to the new characteristics that RAG research is continuously evolving, namely the characteristics of process scheduling and functional componentization. There is currently a lack of comprehensive analysis of the overall RAG system, which has led to research on paradigms lagging behind the development of RAG technology.

\section{Framework and notation}

\begin{table}
    \centering
    \begin{tabular}{cc}
    \toprule
         Notation& Description\\\midrule
         $q$& The original query\\
         $y$& The output of LLM \\
         $D$& A document retrieval repository composed of chunks $d_i$.\\
         $R(q,D)$ & Retriever,find similar chunks from $D$ based on $q$.\\
         $\mathcal F$ & RAG Flow\\
 $\mathcal P$&RAG Flow pattern\\
         $f_{qe}$ &Query expansion function \\
         $f_{qc}$&Query transform function\\
 $f_{comp}$&Chunk compression function\\
 $f_{sel}$&Chunk selection function\\
         $f_r$&Routing function\\
         $M$&Module in modular RAG \\
         $op$&The specific operators within the Module.\\
    \bottomrule
    \end{tabular}
    \caption{Important Notation }
    \label{tab:Notation}
\end{table}

For query $Q= \{q_i\}$, a typical RAG system mainly consists of three key components. \textbf{1) Indexing}. Given documents $\mathcal D=\{d_1,d_2,\ldots,d_n\}$ , where $d_i$ represents the document chunk. Indexing is the process of converting $d_i$ into vectors through an embedding model $f_e(\cdot)$ , and then store vectors in vector database.
\begin{equation}
    \mathcal{I}=\{e_1,e_2,\ldots,e_n\} \quad  and \quad e_i=f_e(d_i)\in \mathbb R^d
\end{equation}\textbf{2) Retrieval }. Transform the query into a vector using the same encoding model, and then filter out the top $k$ document chunks that are most similar based on vector similarity.
\begin{equation}
    \mathcal{R} : \mathop{\mathrm{topk}}_{d_i \in D} \operatorname{Sim}(q, d_i) \rightarrow D^q
\end{equation}
$D^q=\{d_1,d_2,\ldots,d_k\}$ represents the  relevant documents  for question $q$. The similarity function $Sim(\cdot)$  commonly used are dot product or cosine similarity.
\begin{equation}
   Sim(q,d_i)=e_q\cdot e_{d_i} \quad or \quad \frac{e_q \cdot e_{d_i}}{\|e_q\| \cdot \|e_{d_i}\|} 
\end{equation}
\textbf{3) Generation}. After getting the relevant documents. The query $q$ and the retrieved document $D^q$ chunks are inputted together to the LLM to generate the final answer, where $[\cdot , \cdot]$ stands for concatenation.
\begin{equation}
y =\mathcal{LLM}([D^q,q])
\end{equation}
With the evolution of RAG technology, more and more functional components are being integrated into systems.  Modular RAG  paradigm includes three levels, ranging from large to small:

\textbf{L1 Module} ($\mathcal M = \{ \mathcal M_s \}$). The core process in  RAG system.

\textbf{L2 Sub-module} ($\mathcal M_s = \{ Op\}$).The functional modules in module.

\textbf{L3 Operator} ($\mathcal Op = \{f_{\theta_i} \}$). The the specific functional implementation in a module or sub-module.
As a result, a Modular RAG system can be represented as:
\begin{equation}
\mathcal G = \{q,D,\mathcal M,\{\mathcal M_s\},\{Op\}\}
\end{equation}
The arrangement between modules and operators constitutes the RAG Flow $\mathcal F =(M_{\phi_1},\ldots,M_{\phi_n})$
where $\phi $  stands for the set of module parameters. A modular rag flow can be decomposed into a graph of sub-functions. In the simplest case,the graph is a linear chain.
\begin{equation}
Naive RAG :q \xrightarrow[Text-Embedding]{R(q,D)}D^q\xrightarrow[OpenAI/GPT-4]{LLM([q,D^q])}y
\end{equation}
\section{Module and Operator}
This chapter will specifically introduce modules and operators under the Modular RAG framework. Based on the current stage of RAG development, we have established six main modules: Indexing, Pre-retrieval, Retrieval, Post-retrieval, Generation, and Orchestration.

\subsection{Indexing}
Indexing is the process of split document into manageable chunks and it is a key step in organizing a system. Indexing faces three main challenges.
\textbf{1) Incomplete content representation}.The semantic information of chunks is influenced by the segmentation method, resulting in the loss or submergence of important information within longer contexts.
\textbf{2) Inaccurate chunk similarity search}. As data volume increases, noise in retrieval grows, leading to frequent matching with erroneous data, making the retrieval system fragile and unreliable.
\textbf{3) Unclear reference trajectory.} The retrieved chunks may originate from any document, devoid of citation trails, potentially resulting in the presence of chunks from multiple different documents that, despite being semantically similar, contain content on entirely different topics.

\subsubsection{Chunk Optimization}
The size of the chunks and the overlap between the chunks play a crucial role in the overall effectiveness of the RAG system. Given a chunk $d_i$, its chunk size is denoted as $L_i = |d_i|$, and the overlap is denoted as $L^o_i = |d_i \cap d_{i+1}|$. Larger chunks can capture more context, but they also generate more noise, requiring longer processing time and higher costs. While smaller chunks may not fully convey the necessary context, they do have less noise~\cite{searchRAG}.

\textbf{Sliding Window} using overlapping chunks in a sliding window enhances semantic transitions. However, it has limitations such as imprecise context size control, potential truncation of words or sentences, and lacking semantic considerations.


\textbf{Metadata Attachment.} Chunks can be enriched with metadata like page number, file name, author, timestamp, summary, or relevant questions. This metadata allows for filtered retrieval, narrowing the search scope.

\textbf{Small-to-Big }\cite{small2big} separate the chunks used for retrieval from those used for synthesis. Smaller chunks enhance retrieval accuracy, while larger chunks provide more context. One approach is to retrieve smaller summarized chunks and reference their parent larger chunks. Alternatively, individual sentences could be retrieved along with their surrounding text.

\subsubsection{Structure Organization}
One effective method for enhancing information retrieval is to establish a hierarchical structure for the documents. By constructing chunks structure, RAG system can expedite the retrieval and processing of pertinent data.

\textbf{Hierarchical Index}. In the hierarchical structure of documents, nodes are arranged in parent-child relationships, with chunks linked to them. Data summaries are stored at each node, aiding in the swift traversal of data and assisting the RAG system in determining which chunks to extract. This approach can also mitigate the illusion caused by chunk extraction issues. The methods for constructing a structured index primarily include: 1) Structural awareness based on paragraph and sentence segmentation in docs. 2) Content awareness based on inherent structure in PDF, HTML, and Latex. 3) Semantic awareness based on semantic recognition and segmentation of text.

\textbf{KG Index}~\cite{KGP}. Using Knowledge Graphs (KGs) to structure documents helps maintain consistency by clarifying connections between concepts and entities, reducing the risk of mismatch errors. KGs also transform information retrieval into instructions intelligible to language models, improving retrieval accuracy and enabling contextually coherent responses. This enhances the overall efficiency of the RAG system. For example, organizing a corpus in the format of graph $G=\{\mathcal V, \mathcal E, \mathcal X \}$, where node $V =\{v_i\}^n_{i=1}$ represent document structures (e.g.passage, pages, table) , edge  $\mathcal E \subset \mathcal V \times \mathcal V$  represent semantic or lexical similarity and belonging relations, and node features $\mathcal X=\{\mathcal X_i \}^n_{i=1}$ represent text or markdown content for passage.

\subsection{Pre-retrieval}
One of the primary challenges with Naive RAG is its direct reliance on the user’s original query as the basis for retrieval. Formulating a precise and clear question is difficult, and imprudent queries result in subpar retrieval effectiveness. The primary challenges in this module include: \textbf{1) Poorly worded queries}. The question itself is complex, and the language is not well-organized. \textbf{2) Language complexity and ambiguity.} Language models often struggle when dealing with specialized vocabulary or ambiguous abbreviations with multiple meanings. For instance, they may not discern whether \textit{LLM} refers to \textit{Large Language Model} or a \textit{Master of Laws} in a legal context.

\subsubsection{Query Expansion }
Expanding a single query into multiple queries enriches the content of the query, providing further context to address any lack of specific nuances, thereby ensuring the optimal relevance of the generated answers.
\begin{equation}
    f_{qe}(q)=\{q_1, q_2, \ldots, q_n\} \quad \forall q_i \in \{q_1, q_2, \ldots, q_n\}, q_i \notin Q
\end{equation}

\textbf{Multi-Query} uses prompt engineering to expand queries via LLMs, allowing for parallel execution. These expansions are meticulously designed to ensure diversity and coverage. However, this approach can dilute the user's original intent. To mitigate this, the model can be instructed to assign greater weight to the original query.

\textbf{Sub-Query}. By decomposing and planning for complex problems, multiple sub-problems are generated. Specifically, least-to-most prompting~\cite{least-to-most} can be employed to decompose the complex problem into a series of simpler sub-problems. Depending on the structure of the original problem, the generated sub-problems can be executed in parallel or sequentially. Another approach  involves the use of the Chain-of-Verification (CoVe)~\cite{cove}. The expanded queries undergo validation by LLM to achieve the effect of reducing hallucinations. 

\subsubsection{Query Transformation}
Retrieve and generate based on a transformed query instead of the user’s original query.
\begin{equation}
 f_{qt}(q)=q'
\end{equation}

\textbf{Rewrite}. Original queries often fall short for retrieval in real-world scenarios. To address this, LLMs can be prompted to rewrite. Specialized smaller models can also be employed for this purpose~\cite{RRR}. The implementation of the query rewrite method in Taobao has significantly improved recall effectiveness for long-tail queries, leading to an increase in GMV~\cite{BEQUE}.

\textbf{HyDE~\cite{HyDE}.} In order to bridge the semantic gap between questions and answers, it constructs hypothetical documents (assumed answers) when responding to queries instead of directly searching the query. It focuses on embedding similarity from answer to answer rather than seeking embedding similarity for the problem or query. In addition, it also includes reverse HyDE, which generate hypothetical query for each chunks and focuses on retrieval from query to query.

\textbf{Step-back Prompting~\cite{StepBack-prompt}}. The original query is abstracted into a high-level concept question (step-back question). In the RAG system, both the step-back question and the original query are used for retrieval, and their results are combined to generate the language model's answer.

\subsubsection{Query Construction}
In addition to text data, an increasing amount of structured data, such as tables and graph data, is being integrated into RAG systems. To accommodate various data types, it is necessary to restructure the user's query. This involve converting the query into another query language to access alternative data sources, with common methods including Text-to-SQL or Text-to-Cypher . In many scenarios, structured query languages (e.g., SQL, Cypher) are often used in conjunction with semantic information and metadata to construct more complex queries.
\begin{equation}
f_{qc}(q)= q^*,q^* \in Q^*=\{SQL,Cypher,\dots\}
\end{equation}
\subsection{Retrieval}
The retrieval process is pivotal in RAG systems. By leveraging powerful embedding models, queries and text can be efficiently represented in latent spaces, which facilitates the establishment of semantic similarity between questions and documents, thereby enhancing retrieval. Three main considerations that need to be addressed include retrieval efficiency, quality, and the alignment of tasks, data and models.

\subsubsection{\textbf{Retriever Selection}}
With the widespread adoption of RAG technology, the development of embedding models has been in full swing. In addition to traditional models based on statistics and pre-trained models based on the encoder structure, embedding models fine-tuned on LLMs have also demonstrated powerful capabilities~\cite{metb_recent}. However, they often come with more parameters, leading to weaker inference and retrieval efficiency. Therefore, it is crucial to select the appropriate retriever based on different task scenarios.

\textbf{Sparse Retriever} uses statistical methods to convert queries and documents into sparse vectors. Its advantage lies in its efficiency in handling large datasets, focusing only on non-zero elements. However, it may be less effective than dense vectors in capturing complex semantics. Common methods include TF-IDF and BM25.

\textbf{Dense Retriever} employs pre-trained language models (PLMs) to provide dense representations of queries and documents. Despite higher computational and storage costs, it offers more complex semantic representations. Typical models include BERT structure PLMs, like ColBERT, and multi-task fine-tuned models like BGE~\cite{BGE} and GTE~\cite{GTE}.

\textbf{Hybrid Retriever} is to use both sparse and dense retrievers simultaneously. Two embedding techniques complement each other to enhance retrieval effectiveness. Sparse retriever can provide initial screening results. Additionally, sparse models enhance the zero-shot retrieval capabilities of dense models, particularly in handling queries with rare entities, thereby increasing system robustness.

\subsubsection{Retriever Fine-tuning}
In cases where the context may diverge from \ pre-trained corpus, particularly in highly specialized fields like healthcare, law, and other domains abundant in proprietary terminology. While this adjustment demands additional effort, it can substantially enhance retrieval efficiency and domain alignment.

\textbf{Supervised Fine-Tuning (SFT).} Fine-tuning a retrieval model based on labeled domain data is typically done using contrastive learning. This involves reducing the distance between positive samples while increasing the distance between negative samples. The commonly used loss calculation is shown in the following:
\begin{equation}
\mathcal L(\mathcal D_R)= -\frac{1}{T} \sum_{i=1}^{T} \log \frac{e^{(sim(q_i, d_i^+))}}{e^{(sim(q_i, d_i^+))} + \sum_{j=1}^N e^{(sim(q_i, d_i^-))}}
\end{equation}
where  $d_i^+$ is the positive sample document corresponding to the i-th query, $d_i^-$ is several negative sample, $T$ is the total number of queries, $N$  is the number of negative samples, and  $\mathcal D_R$ is the fine-tuning dataset.

\textbf{LM-supervised Retriever (LSR)}. In contrast to directly constructing a fine-tuning dataset from the dataset, LSR utilizes the LM-generated results as supervisory signals to fine-tune the embedding model during the RAG process.
\begin{equation}
P_{LSR}(d|q,y)=\frac{e^{P_{LM}(y |d,q)/\beta}}{\sum_{d' \in D}{e^{P_{LM}(y|d,q)/ \beta)}}}
\end{equation}
$P_{LM}(y|d,q)$  is LM probability of the ground truth output $y$ given the input context $d$ and query $q$, and $\beta$ is a hyperparamter.

\textbf{Adapter.} At times, fine-tuning a large retriever can be costly, especially when dealing with retrievers based on LLMs like gte-Qwen. In such cases, it can mitigate this by incorporating an adapter module and conducting fine-tuning. Another benefit of adding an adapter is the ability to achieve better alignment with specific downstream tasks~\cite{PRCA}.

\subsection{Post-retrieval}
Feeding all retrieved chunks directly into the LLM is not an optimal choice. Post-processing the chunks can aid in better leveraging the contextual information. The primary challenges include:
\textbf{1) Lost in the middle}. Like humans, LLM tends to remember only the beginning or the end of long texts, while forgetting the middle portion~\cite{lostinthemiddle}.
\textbf{2) Noise/anti-fact chunks}. Retrieved noisy or factually contradictory documents can impact the final retrieval generation~\cite{CRUD}.~
\textbf{3) Context Window}. Despite retrieving a substantial amount of relevant content, the limitation on the length of contextual information in large models prevents the inclusion of all this content.

\subsubsection{Rerank}
Rerank the retrieved  chunks without altering their content or length, to enhance the visibility of the more crucial document chunks. Given the retrieved set $D^q$ and a re-ranking method $f_{rerank}$ to obtain the re-ranked set:
\begin{multline}
     D^{q}_r = f_{rerank}(q, D^q) = \{d_1', d_2', \ldots, d_k'\}  \\
     \text{where} f(d'_1) \geq f(d'_2) \geq \ldots \geq f(d'_k).
\end{multline}

\textbf{Rule-base rerank.} Metrics are calculated to rerank chunks according to certain rules. Common metrics include: diversity, relevance and MRR (Maximal Marginal Relevance)~\cite{MRR}. The idea is to reduce redundancy and increase result diversity. MMR selects phrases for the final key phrase list based on a combined criterion of query relevance and information novelty.

\textbf{Model-base rerank.} Utilize a language model to reorder the document chunks, commonly based on the relevance between the chunks and the query. Rerank models have become an important component of RAG systems, and relevant model technologies are also being iteratively upgraded. The scope reordering has also been extended to multimodal data such as tables and images~\cite{Cohere_rerank}.

\subsubsection{Compression}
A common misconception in the RAG process is the belief that retrieving as many relevant documents as possible and concatenating them to form a lengthy retrieval prompt is beneficial. However, excessive context can introduce more noise, diminishing the LLM’s perception of key information. A common approach to address this is to compress and select the retrieved content.
\begin{equation}
    D^q_c = f_{comp}(q, D^q), \quad \text{where} |d_i^{q_c}| < |d_i^q| \quad \forall d_i^q \in D^q
\end{equation}

\textit{\textbf{(Long)LLMLingua}}~\cite{longllmlingua}. By utilizing aligned and trained small language models, such as GPT-2 Small or LLaMA-7B, the detection and removal of unimportant tokens from the prompt is achieved, transforming it into a form that is challenging for humans to comprehend but well understood by LLMs. This approach presents a direct and practical method for prompt compression, eliminating the need for additional training of LLMs while balancing language integrity and compression ratio.

\subsubsection{Selection}
Unlike compressing the content of document chunks, Selection directly removes irrelevant chunks.
\begin{equation}
D^{q}_s = f_{sel}(D^q) = \{ d_i \in D \mid \neg P(d_i) \} 
\end{equation}
Where  $f_{sel}$ is the function for deletion operation and $P(d_i)$ is a conditional predicate indicating that document ($d_i)$ satisfies a certain condition. If document ($d_i$) satisfies ($P(d_i$)), it will be deleted. Conversely, documents for which ($\neg P(d_i)$) is true will be retained.

\textbf{Selective Context}. By identifying and removing redundant content in the input context, the input is refined, thus improving the language model’s reasoning efficiency. In practice, selective context assesses the information content of lexical units based on the self-information computed by the base language model. By retaining content with higher self-information, this method offers a more concise and efficient textual representation, without compromising their performance across diverse applications. However, it overlooks the interdependence between compressed content and the alignment between the targeted language model and the small language model utilized for prompting compression~\cite{Selective-Context}.

\textbf{LLM-Critique}. Another straightforward and effective approach involves having the LLM evaluate the retrieved content before generating the final answer. This allows the LLM to filter out documents with poor relevance through LLM critique. For instance, in Chatlaw~\cite{chatlaw}, the LLM is prompted to self-suggestion on the referenced legal provisions to assess their relevance.

\subsection{Generation}
Utilize the LLM to generate answers based on the user’s query and the retrieved contextual information. Select an appropriate model based on the task requirements, considering factors such as the need for fine-tuning, inference efficiency, and privacy protection.

\subsubsection{Generator Fine-tuning}
In addition to direct LLM usage, targeted fine-tuning based on the scenario and data characteristics can yield better results. This is also one of the greatest advantages of using an on-premise setup LLMs. 

\textbf{Instruct-Tuning}. When LLMs lack data in a specific domain, additional knowledge can be provided to the LLM through fine-tuning. General fine-tuning dataset can also be used as an initial step. Another benefit of fine-tuning is the ability to adjust the model’s input and output. For example, it can enable LLM to adapt to specific data formats and generate responses in a particular style as instructed~\cite{Toolformer}.

\textbf{Reinforcement learning}. Aligning LLM outputs with human or retriever preferences through reinforcement learning is a potential approach~\cite{chatgpt}. For instance, manually annotating the final generated answers and then providing feedback through reinforcement learning. In addition to aligning with human preferences, it is also possible to align with the preferences of fine-tuned models and retrievers.

\textbf{Dual Fine-tuing} Fine-tuning both generator and retriever simultaneously to align their preferences. A typical approach, such as RA-DIT~\cite{RA-DIT}, aligns the scoring functions between retriever and generator using KL divergence. Retrieval likelihood of each retrieved document d is calculated as :
\begin{equation}
    P_R(d|q) = \frac{e^{(sim(d,q))}/\gamma}{\sum_{d\in \mathcal Dq} e^{(sim(d,q)/\gamma}}
\end{equation}
$P_{LM}(y|d,q)$   is the LM probability of the ground truth output y given the input context $d$, question $q$, and $\gamma$ is a hyperparamter. The overall loss is calculated as:
\begin{equation}
    \mathcal L = \frac{1}{|T|} \sum_{i=1}^T KL(P_R(d|q) || P_{LSR}(d|q,y|))
\end{equation}
\subsubsection{Verification }
Although RAG enhances the reliability of LLM-generated answers, in many scenarios, it requires to minimize the probability of hallucinations. Therefore, it can filter out responses that do not meet the required standards through additional verification module. Common verification methods include  knowledge-base and model-base .
\begin{equation}
    y^k = f_{verify}(q,D^q,y)
\end{equation}
\textbf{
Knowledge-base verification} refers to directly validating the responses generated by LLMs through external knowledge. Generally, it extracts specific statements or triplets from response first. Then, relevant evidence is retrieved from verified knowledge base such as Wikipedia or specific knowledge graphs. Finally, each statement is incrementally compared with the evidence to determine whether the statement is supported, refuted, or if there is insufficient information~\cite{wikichat}.

\textbf{Model-based verification} refers to using a small language model to verify the responses generated by LLMs~\cite{KALMV}. Given the input question, the retrieved knowledge, and the generated answer, a small language model is trained to determine whether the generated answer correctly reflects the retrieved knowledge. This process is framed as a multiple-choice question, where the verifier needs to judge whether the answer reflects \textit{correct answer} . If the generated answer does not correctly reflect the retrieved knowledge, the answer can be iteratively regenerated until the verifier confirms that the answer is correct.

\subsection{Orchestration}
Orchestration pertains to the control modules that govern the RAG process. Unlike the traditional, rigid approach of a fixed process, RAG now incorporates decision-making at pivotal junctures and dynamically selects subsequent steps contingent upon the previous outcomes. This adaptive and modular capability is a hallmark of  modular RAG, distinguishing it from the more simplistic Naive and Advance RAG paradigm.

\subsubsection{Routing}
In response to diverse queries, the RAG system routes to specific pipelines tailored for different scenario, a feature essential for a versatile RAG architecture designed to handle a wide array of situations. A decision-making mechanism is necessary to ascertain which modules will be engaged, based on the input from the model or supplementary metadata. Different routes are employed for distinct prompts or components.
This routing mechanism is executed through a function, denoted as \(f_r(\cdot)\), which assigns a score \(\alpha_i\) to each module. These scores dictate the selection of the active subset of modules. Mathematically, the routing function is represented as:
\begin{equation}
    f_r: Q \rightarrow \mathcal{F}
\end{equation}
where \(f_r(\cdot)\) maps the identified query to its corresponding RAG flow.

\textbf{Metadata routing} involves extracting key terms, or entities, from the query, applying a filtration process that uses these keywords and associated metadata within the chunks to refine the  routing parameters. For a specific RAG flow, denoted as \( F_i \), the pre-defined routing keywords are represented as the set \( \mathcal{K}_i = \{k_{i1}, k_{i2}, \ldots, k_{in}\} \). The keyword identified within the query \( q_i \) is designated as \( K'_i \). The matching process for the query \( q \) is quantified by the key score equation:
\begin{equation}
    \text{score}_{\text{key}}(q_i, F_j) = \frac{1}{|K'_j|} | \mathcal{K}_i \cap K'_j |
\end{equation}
This equation calculates the overlap between the pre-defined keywords and those identified in the query, normalized by the count of keywords in \( K'_j \). The final step is to determine the most relevant flow for the query \( q \):
\begin{equation}
    F_i(q) = \text{argmax}_{F_j \in \mathcal{F}} \text{score}(q, F_j) 
\end{equation}
\textbf{Semantic routing} routes to different modules based on the semantic information of the query. Given a pre-defined intent $ \Theta=\{\theta_1,\theta_2,\ldots,\theta_n \}$, the possibility of intent for query q is $P_{\Theta}(\theta|q)=\frac{e^{P_{LM}(\theta |q)}}{\sum_{\theta \in \Theta}{e^{P_{LM}(\theta|q))}}}$. Routing to specific RAG  flow   is determined by the semantic score:

\begin{equation}
   socre_{semantic}(q,F_j)= argmax_{\theta_j\in \Theta }P(\Theta)
\end{equation}
The function $\delta(\cdot)$ serves as a mapping function that assigns an intent to a distinct RAG flow $\mathcal F_i = \delta(\theta_i)$

\textbf{Hybrid Routing} can be implemented to improve query routing by integrating both semantic analysis and metadata-based approaches, which can be defined as follows:
\begin{equation}
   \alpha_i=a \cdot score_{key}(q,F_j) +(1-\alpha)\cdot max_{\theta_j\in \Theta}socre_{semantic}(q,F_j)
\end{equation}
 \( a \) is a weighting factor that balances the contribution of the key-based score and the semantic score.

\subsubsection{Scheduling}
The RAG system evolves in complexity and adaptability, with the ability to manage processes through a sophisticated scheduling module. The scheduling module plays a crucial role in the modular RAG , identifying critical junctures that require external data retrieval, assessing the adequacy of the responses, and deciding on the necessity for further investigation. It is commonly utilized in scenarios that involve recursive, iterative, and adaptive retrieval, ensuring that the system makes informed decisions on when to cease generation or initiate a new retrieval loop.

\textbf{Rule judge.} The subsequent steps are dictated by a set of established rules. Typically, the system evaluates the quality of generated answers through scoring mechanisms. The decision to proceed or halt the process is contingent upon whether these scores surpass certain predetermined thresholds, often related to the confidence levels of individual tokens, which can be defined as follow:
\begin{equation*}
y_t =
\begin{cases}
\hat s_t  \quad \text{    if all tokens of }  \hat s_t \text{ have probs} \ge \tau \\
s_t = LM([D_{q_t},x,y_{<t}])  \quad \text{    otherwise}
\end{cases}
\end{equation*}
Here, \( \hat{s}_t \) represents the tentative answer, and \( s_t \) is the output from the language model. The condition for accepting \( \hat{s}_t \) is that all tokens within it must have associated probabilities greater than or equal to the threshold $\tau$. If this condition is not met, the system reverts to generating a new answer.

\textbf{LLM judge.} The LLM independently determines the subsequent course of action. Two primary approaches facilitate this capability. The first method leverages LLM 's in-context learning capability, and make judgments through prompt engineering. A significant advantage of this method is the elimination of model fine-tuning. Nonetheless, the format of the judgment output is contingent upon the LLM's adherence to the provided instructions.

The second approach involves the LLM generating specific tokens that initiate targeted actions through fine-tuning. This technique, with roots in the Toolformer~\cite{Toolformer}, has been integrated into frameworks like Self-RAG~\cite{self-rag}. This allows for a more direct control mechanism over the LLM's actions, enhancing the system's responsiveness to specific triggers within the conversational context. However, it requires generating a large number of compliant instruction sets to fine-tune LLM.

\textbf{Knowledge-guide scheduling}. Beyond the confines of rule-based methods and the complete reliance on LLMs for process control, a more adaptable intermediate approach emerges with knowledge-guided scheduling~\cite{RoG}. These methods harness the power of  knowledge graphs, to steer the retrieval and generation processes. Specifically, it involves extracting information relevant to the question from a knowledge graph and constructing a reasoning chain. This reasoning chain consists of a series of logically interconnected nodes, each containing critical information for the problem-solving process. Based on the information from the nodes in this reasoning chain, information retrieval and content generation can be performed separately. By integrating this approach, it enhance not only the efficacy and precision of problem-solving but also the clarity of the explanations provided. 

\subsubsection{Fusion}
As RAG process has evolved beyond a linear pipeline, it frequently necessitates broadening the retrieval scope or enhancing diversity by exploring multiple pipelines. Consequently, after the expansion into various branches, the fusion module effectively integrates the information, ensuring a comprehensive and coherent response. The fusion module's reliance is not just for merging answers but also for ensuring that the final output is both rich in content and reflective of the multifaceted nature of the inquiry.

\textbf{LLM fusion}.One of the most straightforward methods for multi-branch aggregation is to leverage the powerful capabilities of LLMs to analyze and integrate information from different branches. However, this approach also faces some challenges, particularly when dealing with long answers that exceeds the LLM's context window limitation. To mitigate this issue, it is common practice to first summarize each branch's answer, extracting the key information before inputting it into the LLM, thus ensuring that the most important content is retained even within length constraints.

\textbf{Weighted ensemble } is based on the weighted values of different tokens generated from multiple branches, leading to the comprehensive selection of the final output.  This approach can be calculated as :
\begin{equation}
    p(y|q,D^q)=\sum_{d\in D^q}p(y|d,q) \cdot \lambda(d,q)
\end{equation}
The weight \( \lambda(d,q) \) is determined by the similarity score between the document \( d \) and the input query \( q \). This weight is calculated using the softmax function, which ensures that the weights are normalized and sum up to one.
\begin{equation}
    \lambda(d,q)=\frac{e^{s(d,q)}}{\sum_{d\in D^q}e^{s(d,q)}}
\end{equation}
\textbf{RRF (Reciprocal Rank Fusion)} is an ensemble technique that synthesizes multiple retrieval result rankings into a cohesive, unified list~\cite{RRF}. It employs a tailored weighted averaging approach to enhance collective predictive performance and ranking precision. The method's strength is its dynamic weight assignment, which is informed by the interplay among branches. RRF is especially potent in scenarios characterized by model or source heterogeneity, where it can markedly amplify the accuracy of predictions.

\section{RAG Flow and Flow Pattern}
The collaboration between operators forms the workflow of the module, which we refer to as RAG flow $\mathcal F =(M_{\phi_1},\ldots,M_{\phi_n})$, where $\phi$   stands for the set of module parameters. A modular rag flow can be decomposed into a graph of sub-functions. Through control logic, the operators can execute in a predetermined pipeline, while also performing conditional, branching or looping when necessary. In the simplest case. the graph is a linear chain.

After conducting an in-depth analysis of current RAG  methods, we have identified a set of common RAG flow patterns, denoted as \(\mathcal{P}\). These patterns transcend various application domains and demonstrate a high level of consistency and reusability, revealing the prevalent structures and behaviors in process design. A RAG flow pattern can be defined as $\mathcal P = \{M_{\phi_1}:\{\mathcal Op_1\}\rightarrow M_{\phi_2}:\{\mathcal Op_2\} \rightarrow \ldots \rightarrow M_{\phi_n}:\{\mathcal Op_n\}\}$

\subsection{Linear Pattern }
\begin{figure}
    \centering
    \includegraphics[width=\linewidth]{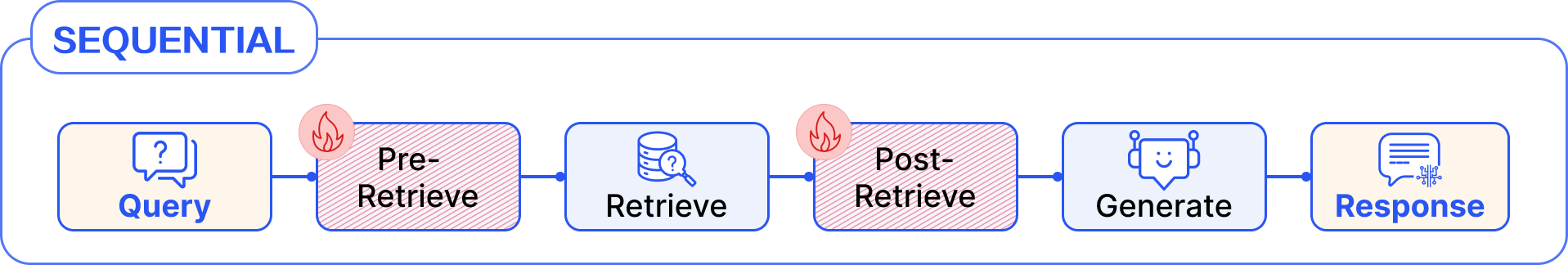}
    \caption{Linear RAG flow pattern. Each module is processed in a fixed sequential order. }
    \label{fig:Linear}
\end{figure}

The modules in the modular RAG system are organized in a linear way, and can be described as Algorithm~\ref{alg:linear}. 
\begin{equation}
    \mathcal P_{linear} =\{M_1 \rightarrow M_2 \rightarrow \ldots \rightarrow M_n \} 
\end{equation}
The linear flow pattern is the simplest and most commonly used pattern. As shown in Figure~\ref{fig:Linear}, the full linear RAG flow pattern mainly includes pre-retrieval processing, retrieval, post-retrieval processing, and generation modules. 
$\mathcal P_{linear_{full}} =\{M_{\text{indexing}} \rightarrow M_{\text{pre-retrieval}} \rightarrow M_{\text{retrieval}} \rightarrow M_{\text{post-retrieval}} \rightarrow M_{\text{generate}} \} $. If there are no pre-retrieval and post-retrieval modules, it follows the Naive RAG paradigm.

\begin{algorithm}
\caption{Linear RAG Flow Pattern}
\label{alg:linear}
\begin{algorithmic}[1]
\REQUIRE original query $q$, documents $D$, retriever $R$, language model $LLM$, pre-processing function $f_{pre}$, post-processing function $f_{post}$
\ENSURE final output $\hat y$
\STATE Initialize:
\STATE $q' \leftarrow f_{pre}(q)$ // Pre-process the original query
\STATE $D^{q'} \leftarrow R(q', D)$ // Retrieve documents related to the pre-processed query
\STATE $\hat D^{q'} \leftarrow f_{post}(q', D^{q'})$ // Post-process the retrieved documents
\STATE $\hat y \leftarrow LLM([q, \hat D^{q'}])$ // Generate output using the language model with the original query and post-processed documents
\RETURN $\hat y$ // Return the final output
\end{algorithmic}
\end{algorithm}

\begin{figure}
    \centering
    \includegraphics[width=\linewidth]{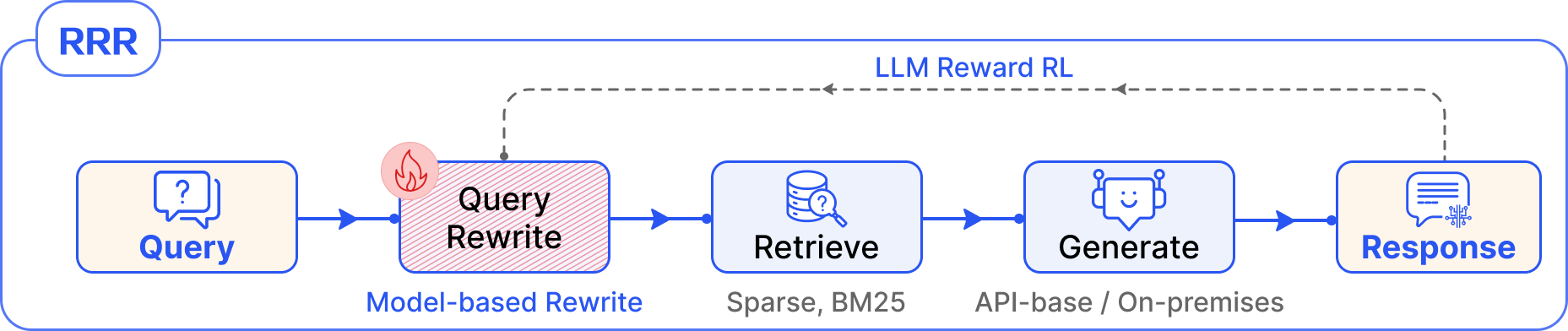}
    \caption{RRR~\cite{RRR} is a typical linear flow that introduces a learnable query rewrite module before retrieval. This module employs reinforcement based on the output results of the LLM.}
    \label{fig:RRR}
\end{figure}

Common linear RAG flow involves a query transform module (such as rewrite or HyDE operators) at the pre-retrieval stage and utilize rerank at the post-retrieval stage. Rewrite-Retrieve-Read (RRR)~\cite{RRR} is  a typical linear structure. As illustrated in Figure~\ref{fig:RRR}, the query rewrite module $f_{rewrite}$ is a smaller trainable language model  fine-tuned on T5-large, and in the context of reinforcement learning, the optimization of the rewriter is formalized as a Markov decision process, with the final output of the LLM serving as the reward. The retriever utilizes a sparse encoding model, BM25.

\subsection{Conditional Pattern}
\begin{figure}
    \centering
    \includegraphics[width=\linewidth]{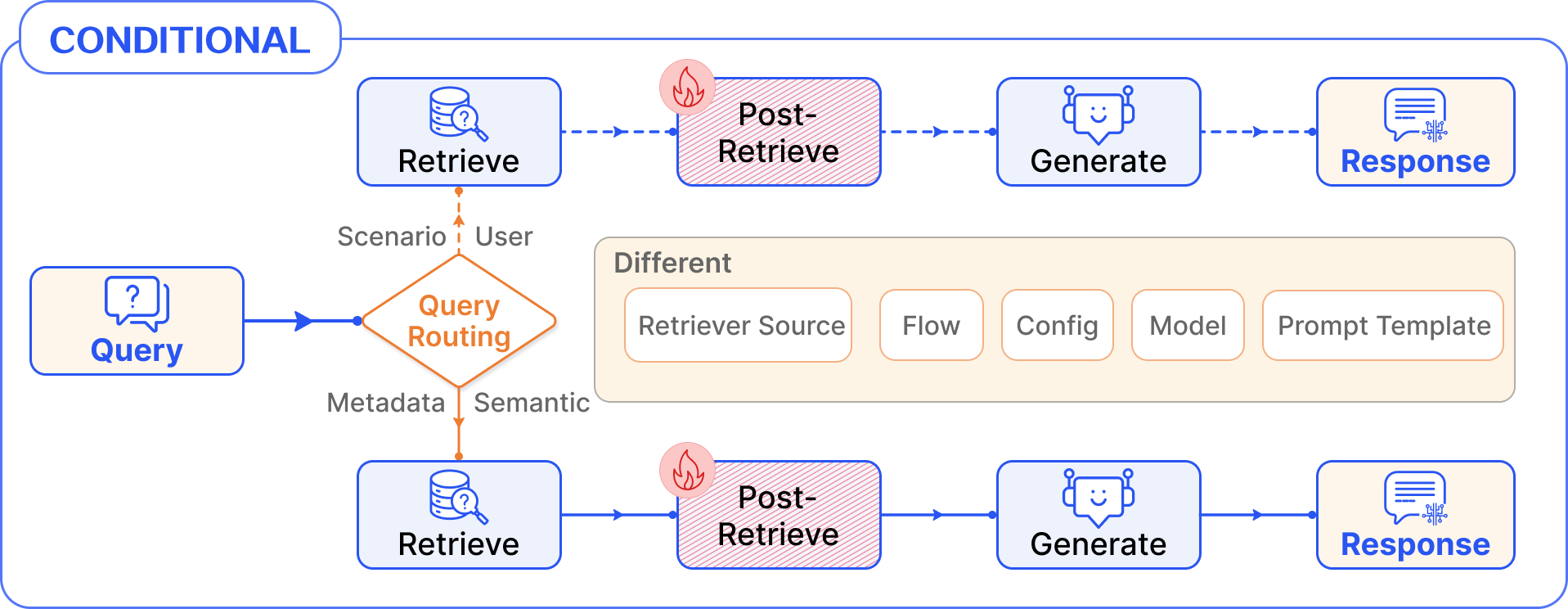}
    \caption{The conditional flow pattern. There is a routing module that controls which RAG flow the query is directed to. Typically, different flows are used for various configurations to meet the general requirements of the RAG system.}
    \label{fig:conditonal}
\end{figure}

The RAG flow with conditional structure involves selecting different RAG pipeline based on different conditions, as illustrated in Figure~\ref{fig:conditonal}. A  detailed definition is shown in Algorithm~\ref{alg:conditional}. Typically, pipleline selection is accomplished through a routing module that determines the next module in the flow.
\begin{equation}
    \mathcal P_{\text{conditional}} = \{ M_i \xrightarrow[]{f_r} M_j \lor M_k \}
\end{equation}
Where $\xrightarrow[]{f_r}$ represents that based on routing function $f_r(\cdot)$, the flow can go to module $M_j$ or $M_k$.

\begin{algorithm}
\caption{Conditional RAG Flow Pattern}
\label{alg:conditional}
\begin{algorithmic}[1]
\REQUIRE original query $q$, documents $D$, language model $LM$, retriever $R$, routing function $f_r$
\ENSURE final output $\hat y$
\STATE Initialize:
\STATE $q' \leftarrow \text{QueryTransform}(q)$ // Pre-process the initial query if needed
\STATE $D' \leftarrow R(q', D)$ // Retrieve or update documents related to the query
\STATE $M_{\text{next}} \leftarrow f_r(q', D')$ // Determine the next module using the routing function
\IF{$M_{\text{next}} = M_j$}
    \STATE $\hat y \leftarrow M_j(q', D')$ // Execute module $M_j$
\ELSIF{$M_{\text{next}} = M_k$}
    \STATE $\hat y \leftarrow M_k(q', D')$  $M_k$
\ENDIF
\RETURN $\hat y$ 
\end{algorithmic}
\end{algorithm}

Pipeline selection is determined by the nature of the question, directing different flows tailored to specific scenarios. For example, the tolerance for responses generated by LLMs varies across questions related to serious issues, political matters, or entertainment topics. These routing flow often diverge in terms of retrieval sources, retrieval processes, configurations, models, and prompts.

\begin{figure}
    \centering
    \includegraphics[width=\linewidth]{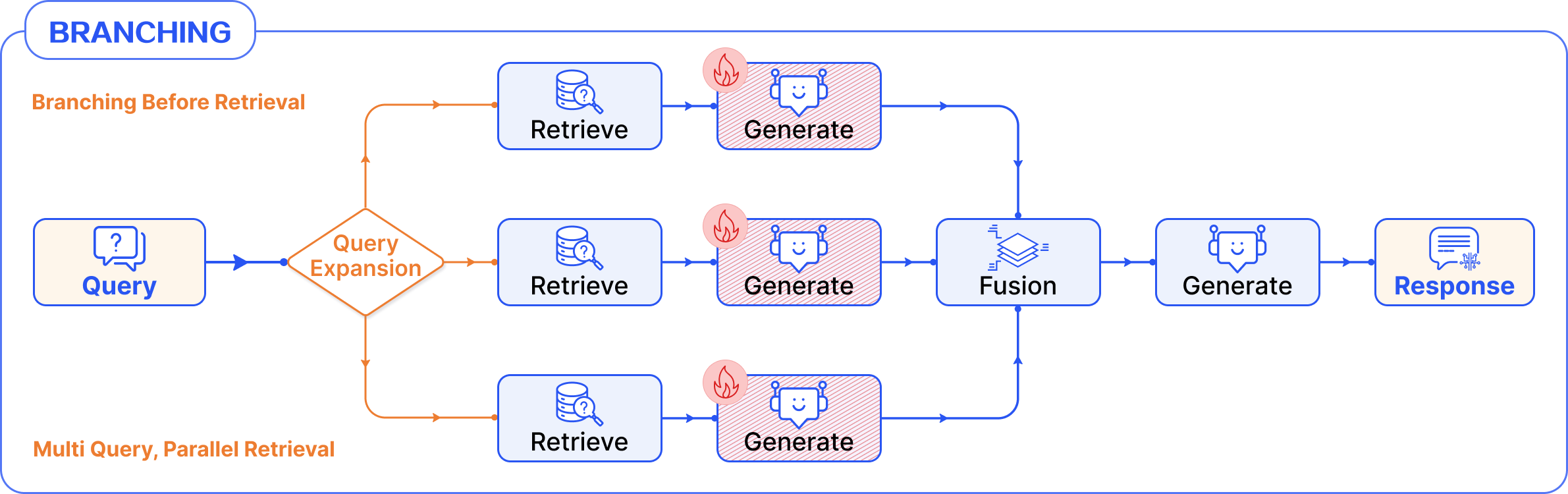}
    \caption{Pre-retrieval branching flow pattern.Each branch performs retrieval and generation separately, and then they are aggregated at the end.}
    \label{fig:pre_branch}
\end{figure}

\begin{figure}
    \centering
    \includegraphics[width=\linewidth]{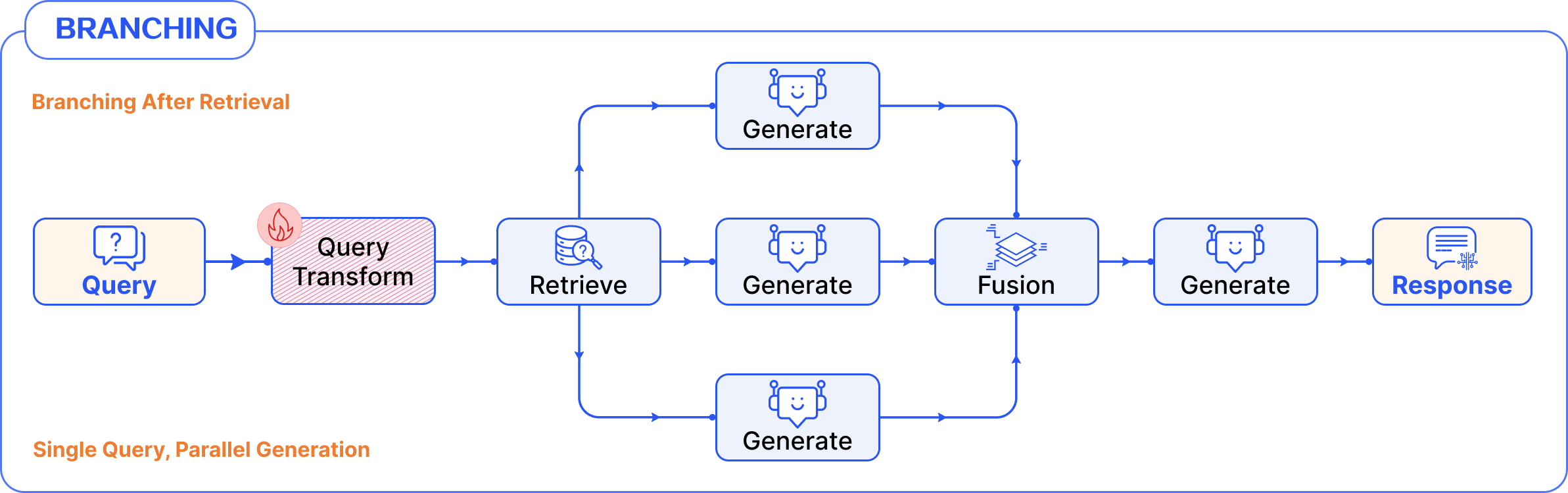}
    \caption{Post-retrieval branching flow pattern.Only one retrieval performed, and then generation is carried out separately for each retrieved document chunks, followed by aggregation.}
    \label{fig:post_branch}
\end{figure}

\subsection{Branching}
In many cases, the RAG flow system may have multiple parallel running branches , usually to increase the diversity of generated results. Assuming multiple branches $b_i$ are generated in module $B = M_{\text{split}}(\cdot) = \{b_1, b_2, \ldots, b_m\}$. For each branch $b_i \in B$, the same or different RAG processes can be executed, passing through multiple processing modules $\{M_{1}, M_{2}, \ldots, M_{k}\}$ to obtain branch output result $p_i = M_{ik}(\ldots M_{i2}(M_{i1}(b_i)) \ldots)$. The results of multiple branches are aggregated using an aggregation function to obtain intermediate output results. $\hat O = M_{\text{merge}}(\{p_i \mid b_i \in B\})$. However, aggregation is not necessarily the end of the RAG flow, as it can continue to connect to other modules, $M_{jn}(\ldots M_{j2}(M_{j1}(\hat O)) \ldots)$. For example, after aggregating multiple model responses, they can continue through a validation module. Therefore, the entire branch flow pattern can be represented as:
\begin{equation}
\begin{split}
    \mathcal P_{branch}= & M_{jn}(\ldots M_{j1}(M_{\text{merge}}( \{M_{ik} \\
    & (\ldots M_{i1}(b_i) \ldots) \mid b_i \in M_{\text{split}}(q)\}))\ldots)
\end{split}
\end{equation}
\begin{algorithm}
\caption{Pre-retrieval Branching Flow Pattern}
\label{alg:pre-retrieval-branch}
\begin{algorithmic}[1]
\REQUIRE original query $q$, documents $D$, query expand module  $M_{\text{expand}}$, retriever $M_{\text{retrieve}}$, language model $LLM$, merge module  $M_{\text{merge}}$
\ENSURE final output $\hat y$
\STATE Initialize:
\STATE $Q' \leftarrow M_{\text{expand}}(q)$ // Expand the original query to multiple sub-queries
\FORALL{$q'_i \in Q'$}
    \STATE $D'_i \leftarrow M_{\text{retrieve}}(q'_i, D)$ // Retrieve documents for each sub-query
    \STATE $G_i \leftarrow \emptyset$ // Initialize an empty set for generated results of the sub-query
    \FORALL{$d'_{ij} \in D'_i$}
        \STATE $y_{ij} \leftarrow LLM([q'_i, d'_{ij}])$ // Generate results for each document of the sub-query
        \STATE $O_i \leftarrow O_i \cup \{y_{ij}\}$ // Add generated results to the set
    \ENDFOR
    \STATE $\hat y \leftarrow M_{\text{merge}}(O_i)$  // Merge generated results of the sub-query into the final result
\ENDFOR
\RETURN $\hat y$ 
\end{algorithmic}
\end{algorithm}

The RAG flow with a branching structure differs from the conditional approach in that it involves multiple parallel branches, as opposed to selecting one branch from multiple options in the conditional approach. Structurally, it can be categorized into two types, which are depicted in Figure~\ref{fig:pre_branch} and Figure~\ref{fig:post_branch}.

\textbf{Pre-Retrieval Branching }(Multi-Query, Parallel Retrieval)\textbf{.}  As shown in Algorithm~\ref{alg:pre-retrieval-branch}, the process involves initially taking a query \( q \) and expanding it through a module \( M_{\text{expand}} \) to generate multiple sub-queries \( Q' \). Each sub-query \( q'_i \) is then used to retrieve relevant documents via \( M_{\text{retrieve}} \), forming document sets \( D'_i \). These document sets, along with the corresponding sub-queries, are fed into a generation module \( M_{\text{generate}} \) to produce a set of answers \( G_i \). Ultimately, all these generated answers are combined using a merging module \( M_{\text{merge}} \) to form the final result \( y \). This entire flow can be mathematically represented as:
\begin{equation}
\begin{split}
\mathcal P_{branch_{pre}} = & M_{\text{merge}} ( _{q'_i \in M_{\text{expand}}(q)} \{ M_{\text{generate}}(q'_i, d'_{ij}) \mid  \\
& d'_{ij} \in M_{\text{retrieve}}(q'_i) \} )
\end{split}
\end{equation}

\textbf{Post-Retrieval Branching} (Single Query, Parallel Generation). As shown in Algorithm~\ref{alg:post-retrieval-branch}, in the post-retrieval branching pattern, the process starts with a single query \( q \) which is used to retrieve multiple document chunks through a retrieval module \( M_{\text{retrieve}} \), resulting in a set of documents \( D^q \). Each document \( d^q_i \) from this set is then independently processed by a generation module \( M_{\text{generate}} \) to produce a set of generated results \( G \). These results are subsequently merged using a merge module \( M_{\text{merge}} \) to form the final result \( y \). The process can be succinctly represented as \( y = M_{\text{merge}}(O_i) \), where \( O_i \) is the collection of all generated results from each document \( d^q_i \) in \( D^q \).
Therefore, the entire process can be represented as:
\begin{equation}
    \mathcal P_{branch_{post}}= M_{\text{merge}}(\{M_{\text{generate}}(d^q_i) \mid d^q_i \in M_{\text{retrieve}}(q)\})
\end{equation}

\begin{algorithm}
\caption{Post-retrieval Branching Flow Pattern}
\label{alg:post-retrieval-branch}
\begin{algorithmic}[1]
\REQUIRE original query $q$, documents $D$, retriever $R$, language model $LLM$, merge module $M_{\text{merge}}$
\ENSURE final output $\hat y$
\STATE Initialize:
\STATE $q' \leftarrow f_{pre}(q)$ // Pre-process the original query
\STATE $D^{q'} \leftarrow R(q', D)$ // Retrieve a set of documents based on the pre-processed query
\STATE $G \leftarrow \emptyset$ // Initialize an empty set to store generated results
\FORALL{$d_i \in D^{q'}$}
    \STATE $y_i \leftarrow LLM([q, d_i])$ // Generate results independently for each document chunk using the language model
    \STATE $O_i \leftarrow O_i \cup \{y_i\}$ // Add the generated result to the set of results
\ENDFOR
\STATE $\hat y \leftarrow M_{\text{merge}}(O_i)$ // Merge all generated results using the merge function
\RETURN $\hat y$ 
\end{algorithmic}
\end{algorithm}

REPLUG~\cite{Replug} embodies a classic post-retrieval branching structure, wherein the probability of each token is predicted for each branch. Through weighted possibility ensemble, the different branches are aggregated, and the final generation result is used to fine-tune the retriever, known as Contriever, through feedback.

\begin{figure}
    \centering
    \includegraphics[width=\linewidth]{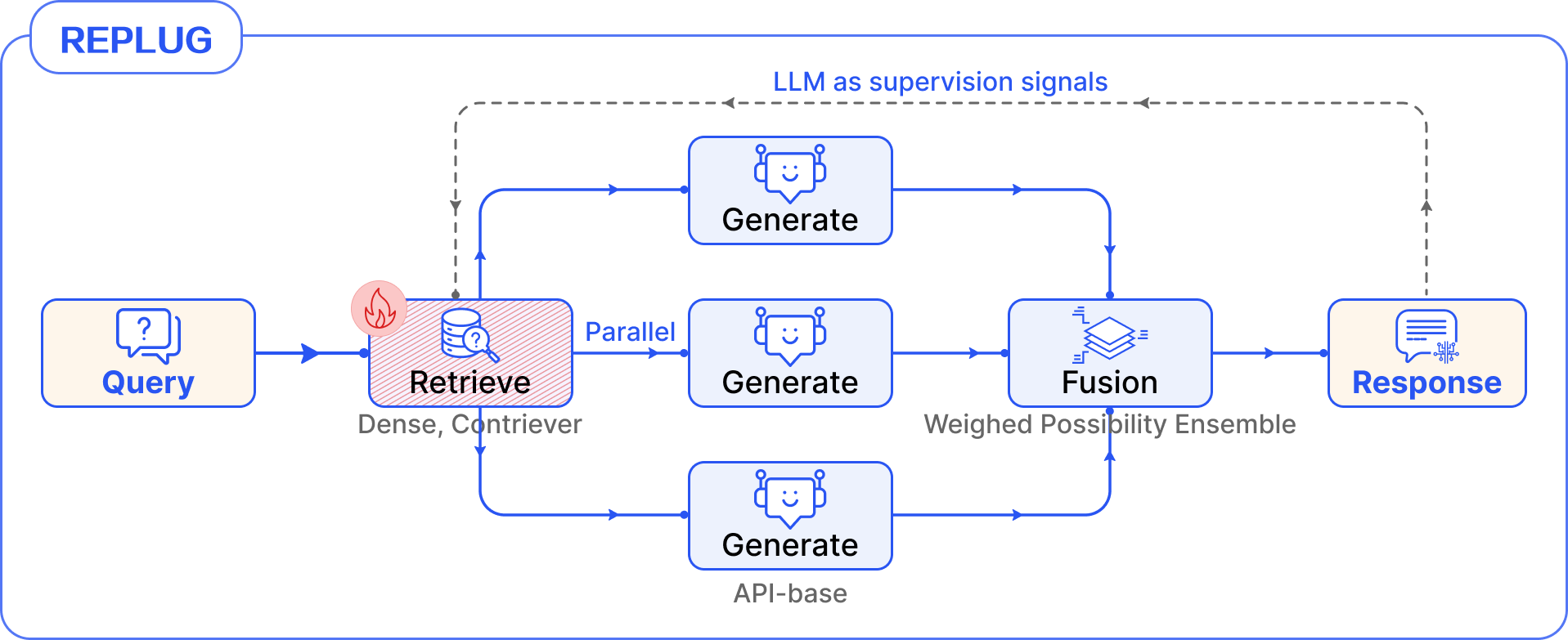}
    \caption{The RAG flow in REPLUG~\cite{Replug}, which follows a typical post-retrieval branching pattern. Each retrieved chunks undergoes parallel generation, and then they are aggregated using a weighted probability ensemble. }
    \label{fig:Replug}
\end{figure}

\subsection{Loop Pattern}
The RAG flow with a loop structure, as an important characteristic of Modular RAG, involves interdependent retrieval and generation steps. It typically includes a scheduling module for flow control. The modular RAG system can be abstracted as a directed graph $G = (V, E)$, where \( V \) is the set of vertices representing the various modules \( M_i \) in the system, and \( E \) is the set of edges representing the control flow or data flow between modules. If there is a vertex sequence \( M_{i_1}, M_{i_2}, ..., M_{i_n} \) such that \( M_{i_n} \) can reach \( M_{i_1} \) (i.e., \( M_{i_n} \rightarrow M_{i_1} \)), then this RAG system forms a loop. If \( M_j \) is the successor module of \( M_i \) and \( M_i \) decides whether to return to \( M_j \) or a previous module \( M_k \) through a Judge module, it can be represented as: $M_i \xrightarrow{\text{Judge}} M_j \quad \text{or} \quad M_i \xrightarrow{\text{Judge}} M_k $ where $M_k$ is the predecessor module of $M_j$. If \( M_i \) return to \( M_j \), it can be represented as:
$\exists \text{Judge}(M_i, M_j) \quad \text{s.t.} \quad (M_i, M_j) \in E \quad $ and $\text{Judge}(M_i, M_j) = \text{true}$. If the Judge module not to return to any previous module, it can be represented as:
$\forall M_i \in V, \quad \text{Judge}(M_i, M_j) = \text{false}$ for all $M_j$ that are predecessors of $M_i$. Loop pattern can be further categorized into iterative, recursive, and adaptive (active) retrieval approaches.
 \begin{figure}
    \centering
    \includegraphics[width=\linewidth]{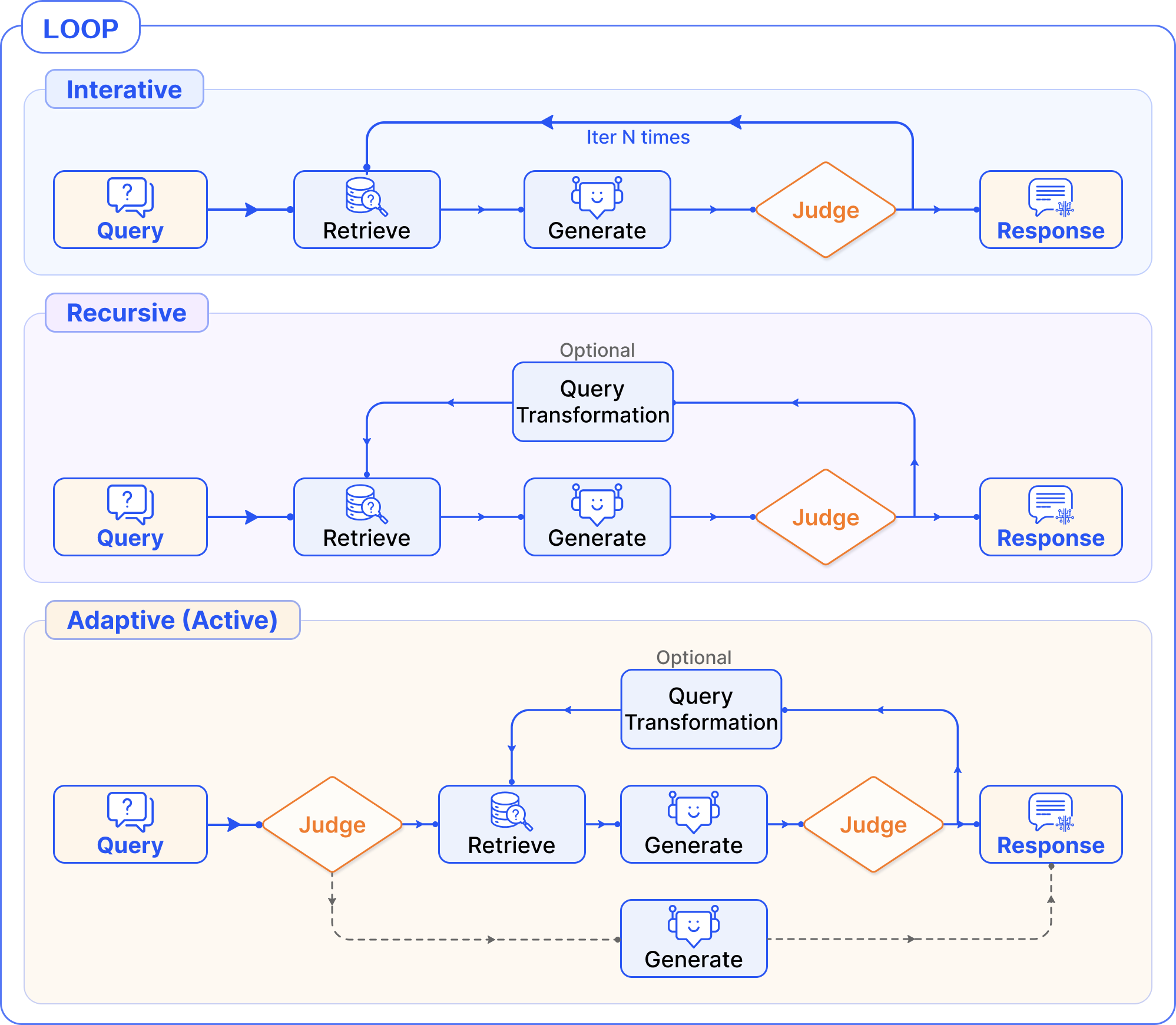}
    \caption{Loop flow pattern. Typically, a RAG system performs multiple rounds of retrieval and generation. It can be categorized into three forms: iterative, recursive, and adaptive.}
    \label{fig:loop}
\end{figure}

\textbf{Iterative retrieval}
At times, a single retrieval and generation may not effectively address complex questions requiring extensive knowledge. Therefore, an iterative approach can be used in RAG (see Algorithm~\ref{alg:iter}), typically involving a fixed number of iterations for retrieval. At step \( t \), given the query \( q_t \) and the previous output sequence $y_{<t}=[y_0,\ldots,y_{t-1}]$ , iterations proceed under the condition that \( t \) is less than the maximum allowed iterations \( T \). In each loop, it retrieves a document chunks  \( D_{t-1} \) using the last output \( y_{t-1} \) and the current query \( q_t \). Subsequently, a new output \( y_t \) is generated. The continuation of the iteration is determined by a Judge module, which makes its decision based on the \( y_t \),  \( y_{<t} \),  \( q_t \), and the \( D_{t-1} \).

\begin{algorithm}
\caption{Iterative RAG Flow Pattern}
\label{alg:iter}
\begin{algorithmic}[1]
\REQUIRE original query $q$, documents $D$, maximum iterative times $T$, language model $LLM$, retriever $R$, initial output $y_{<1} = \emptyset$
\ENSURE final output $\hat y$
\STATE Initialize:
\STATE $q_t \leftarrow q$ // Initialize query for the first iteration
\STATE $y_{<1} \leftarrow \emptyset$ // Initialize previous outputs as empty
\STATE $t \leftarrow 1$ // Initialize iteration step
\WHILE{$t \leq T$}
    \STATE $q_t \leftarrow \text{QueryTransform}(y_{<t-1}, q_{t-1})$ // Generate query based on previous output and original query
    \STATE $D_t \leftarrow R(y_{t-1}||q_t, D)$ // Retrieve or update documents related to the current query
    \STATE $y_t \leftarrow LLM([y_{<t-1}, q_t, D_t])$ // Generate output using the language model
    \STATE $y_{<t} \leftarrow [y_{<t-1}, y_t]$ // Update the list of previous outputs
    \IF{$\text{Judge}(y_t, q) = \text{false}$}
        \STATE break
    \ENDIF
    \STATE $t \leftarrow t + 1$ // Increment iteration step
\ENDWHILE
\STATE $y_{final}=\text{synthesizeOutput}(y_{\leq t})$ // Synthesize final output from the list of outputs
\RETURN $ \hat y$
\end{algorithmic}
\end{algorithm}

 \begin{figure}
    \centering
    \includegraphics[width=\linewidth]{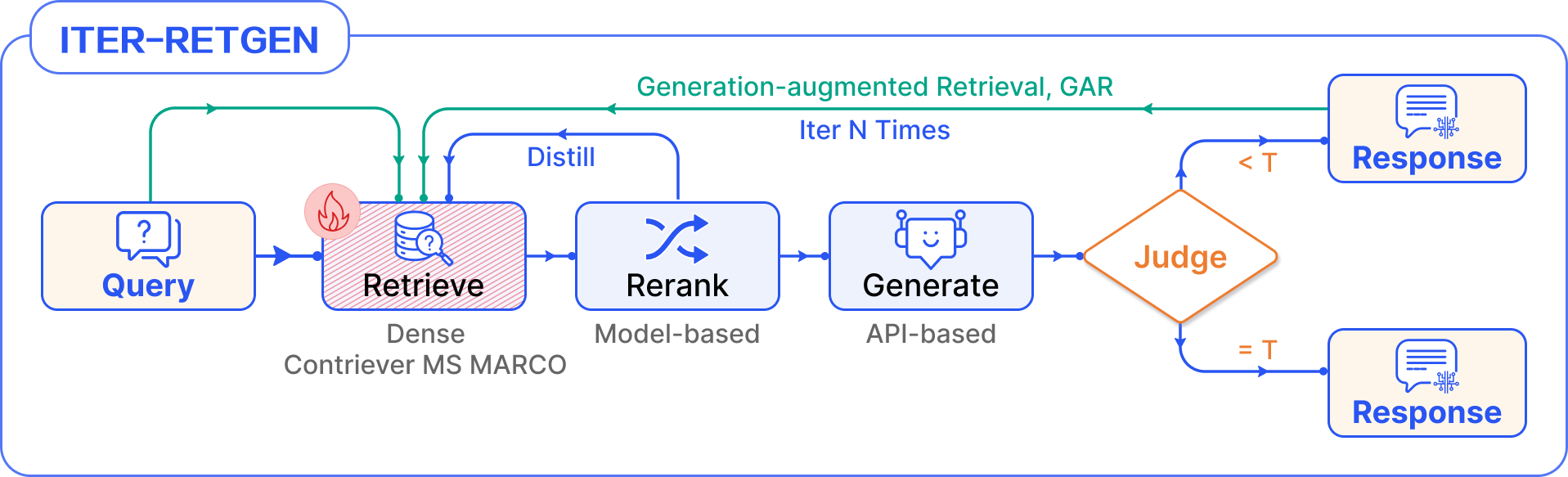}
    \caption{ITER-RETGEN~\cite{ITER-RETGEN} is a typical iterative structure. Multiple rounds of retrieval and generation are performed within the limit of the maximum number of iterations. }
    \label{fig:iter-retgen}
\end{figure}

An exemplary case of iterative retrieval is ITER-RETGEN~\cite{ITER-RETGEN} (Figure~\ref{fig:iter-retgen}), which iterates retrieval-augmented generation and generation-augmented retrieval. Retrieval-augmented generation outputs a response to a task input based on all retrieved knowledge. In each iteration, ITER-RETGEN leverages the model output from the previous iteration as a specific context to help retrieve more relevant knowledge. Termination of the loop is determined by a predefined number of iterations.

\textbf{Recursive retrieval}
The characteristic feature of recursive retrieval (see Algorithm~\ref{alg:recursive}), as opposed to iterative retrieval, is its clear dependency on the previous step and its continuous deepening of retrieval. Typically, it follows a tree-like structure and there is a clear termination mechanism as an exit condition for recursive retrieval. In RAG systems, recursive retrieval usually involves query transform, relying on the newly rewritten query for each retrieval. 

\begin{algorithm}
\caption{Recursive RAG Flow Pattern}
\label{alg:recursive}
\begin{algorithmic}[1]
\REQUIRE initial query $q$, document $D$, retriever $R$, language model $LM$, maximum recursive depth $K_{max}$
\ENSURE final output $\hat y$
\STATE Initialize:
\STATE $\quad Q \leftarrow \{q\} $
\STATE $\quad k \leftarrow 0$ // Initialize recursion depth
\WHILE{$Q \neq \emptyset$ \AND $k < K_{max}$}
\STATE $Q' \leftarrow \emptyset$ // To store queries for the next recursion level
\FORALL{$q \in Q$}
\STATE $D^q \leftarrow R(q, D)$ // Retrieve or update documents related to the current query
\STATE $Y \leftarrow LM([q, D^q])$ // Generate outputs using the language model
\STATE $Q'' \leftarrow \text{deriveNewQueries}(q,D^q,Y)$ // Derive new queries from generated outputs
\FORALL{$q' \in Q''$}
\IF{$q' \notin Q'$ \AND $q' \notin Q$}
\STATE $Q' \leftarrow Q' \cup \{q'\}$
\ENDIF
\ENDFOR
\ENDFOR
\STATE $Q \leftarrow Q'$ // Update the set of queries for the next recursion

\STATE $k \leftarrow k + 1$ // Increment recursion depth
\ENDWHILE
\STATE  $\hat y=\text{synthesizeOutput}(Y)$ // Synthesize final output from generated outputs
\RETURN $\hat y$
\end{algorithmic}
\end{algorithm}

\begin{figure}
    \centering
    \includegraphics[width=\linewidth]{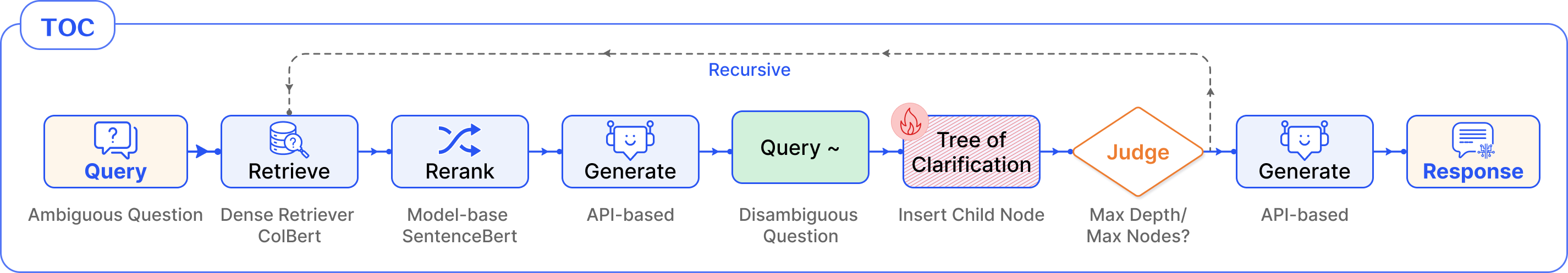}
    \caption{RAG flow of ToC~\cite{TOC}. A typical characteristic of this process is that each recursive retrieval uses the new query generated from the previous step, thereby progressively deepening analysis of the original complex query.}
    \label{fig:toc}
\end{figure}

A typical implementation of recursive retrieval, such as ToC~\cite{TOC} (see Figure~\ref{fig:toc} ), involves recursively executing RAC (Recursive Augmented Clarification) to gradually insert sub-nodes into the clarification tree from the initial ambiguous question (AQ). At each expansion step, paragraph re-ranking is performed based on the current query to generate a disambiguous Question (DQ). The exploration of the tree concludes upon reaching the maximum number of valid nodes or the maximum depth. Once the clarification tree is constructed, ToC gathers all valid nodes and generates a comprehensive long-text answer to address AQ.

\textbf{Adaptive (Active) retrieval}
With the advancement of RAG, there has been a gradual shift beyond passive retrieval to the emergence of adaptive retrieval (see Algorithm~\ref{alg:active}) , also known as active retrieval, which is partly attributed to the powerful capabilities of LLM. This shares a core concept with LLM Agent~\cite{metagpt}. RAG systems can actively determine the timing of retrieval and decide when to conclude the entire process and produce the final result. Based on the criteria for judgment, this can be further categorized into Prompt-base and Tuning-base approaches.

\begin{figure}
    \centering
    \includegraphics[width=\linewidth]{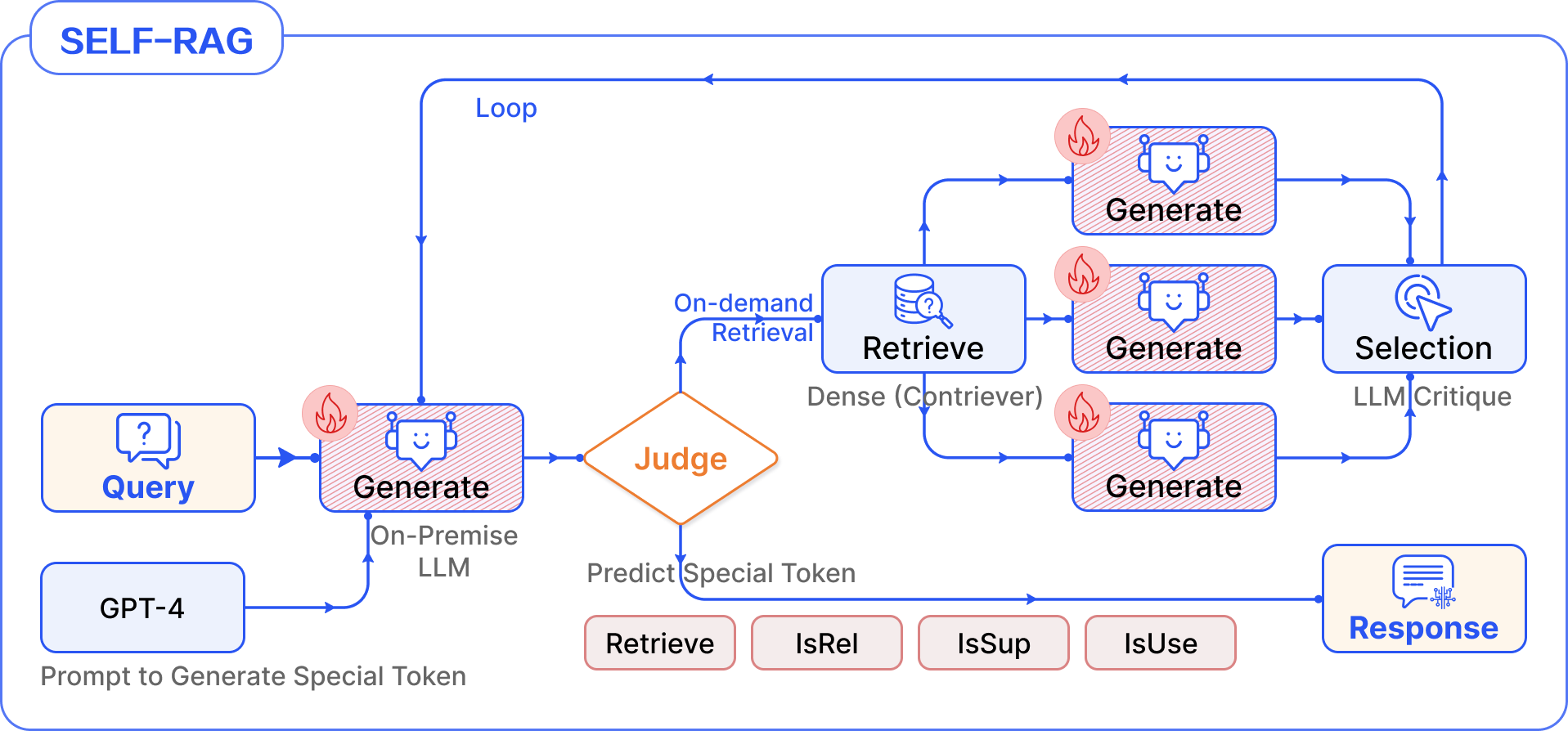}
    \caption{RAG flow of SELF-RAG~\cite{self-rag}. First, it prompt GPT-4 to obtain a suitable instruct fine-tuning dataset to fine-tune the deployed open-source LLM. This allows the model to output four specific tokens during generation, which are used to control the RAG process.}
    \label{fig:self-rag}
\end{figure}

\begin{algorithm}
\caption{Active RAG Flow Pattern}
\label{alg:active}
\begin{algorithmic}[1]
\REQUIRE original query $Q$, documents $D$, maximum iterative times $T$, language model $LLM$, retriever $R$
\ENSURE final output $\hat y$
\STATE Initialize:
\STATE $t \leftarrow 1$ // Initialize loop step
\STATE $q_t \leftarrow q$ // Initialize query for the first iteration
\STATE $y_{<1} \leftarrow \emptyset$ // Initialize previous outputs as empty
\WHILE{$t \leq T$}
    \STATE $Q_{t} \leftarrow \text{QueryTransform}(y_{<t-1}, q_{t-1})$ // Derive new query from previous output and query
    \IF{$\text{Evaluate}(Q_t, y_{<t-1})$}
        \STATE $D_t \leftarrow R(q_t, D)$ // Retrieve documents based on the new query
        \STATE $y_t \leftarrow LLM([q_t, D_t])$ // Generate output using the language model
    \ELSE
        \STATE $y_t \leftarrow \emptyset$ // Set output as empty if query evaluation is false
    \ENDIF
    \STATE $y_{<t} \leftarrow [y_{<t-1}, y_t]$ // Update the list of previous outputs
    \IF{$\text{isOutputAcceptable}(y_t, y_{<t}, q_t) = \text{false}$}
        \STATE break // Break if the output is not acceptable
    \ENDIF
    \STATE $t \leftarrow t + 1$ // Increment iteration step
\ENDWHILE
\STATE $\hat y =\text{synthesizeOutput}(y_{\leq t})$ // Synthesize final output from the list of outputs
\RETURN $\hat y$
\end{algorithmic}
\end{algorithm}
\textbf{Prompt-base.} The prompt-base approach involves controlling the flow using Prompt Engineering to direct LLM. A typical implementation example is FLARE~\cite{Flare}. Its core concept is that LLMs should only retrieve when essential knowledge is lacking, to avoid unnecessary or inappropriate retrieval in an enhanced LM. FLARE iteratively generates the next provisional sentence and checks for the presence of low-probability tokens. If found, the system retrieves relevant documents and regenerates the sentence.
 \begin{figure}
    \centering
    \includegraphics[width=\linewidth]{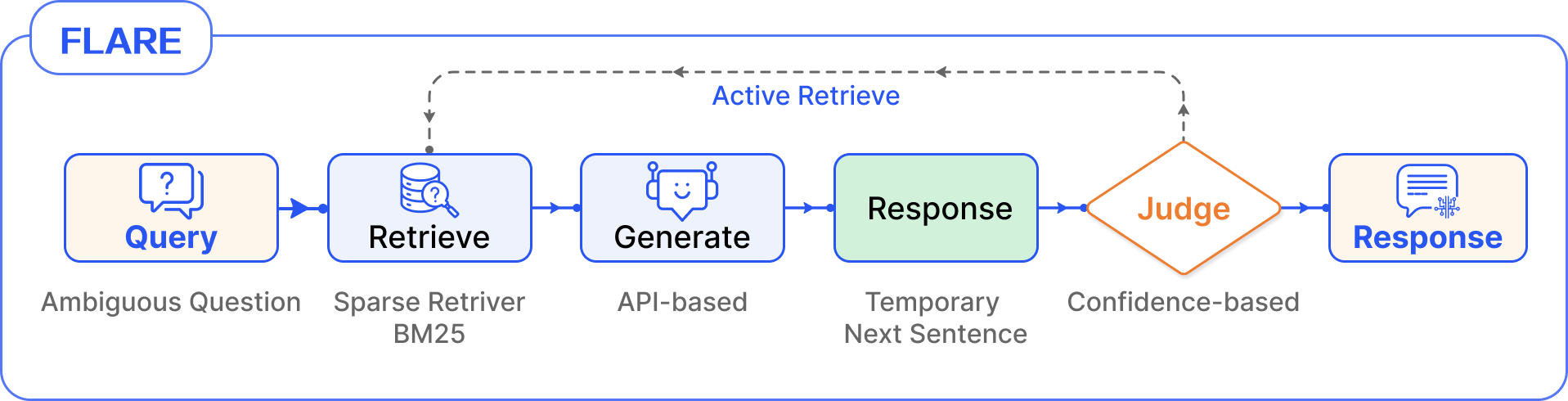}
    \caption{RAG flow of FLARE~\cite{Flare}. The generated provisional answer will undergo confidence assessment. If it does not meet the required confidence level, the process will return to the retrieval stage and generate anew. The assessment criteria are implemented through prompt }
    \label{fig:flare}
\end{figure}
\textbf{Tuning-base. }The tuning-based approach involves fine-tuning LLM to generate special tokens, thereby triggering retrieval or generation. This concept can be traced back to Toolformer~\cite{Toolformer}, where the generation of specific content assists in invoking tools. In RAG systems, this approach is used to control both retrieval and generation steps. A typical case is Self-RAG~\cite{self-rag}(see Figure~\ref{fig:self-rag}). Given an input prompt and the preceding generation result, first predict whether the special token \textit{Retrieve} is helpful for enhancing the continued generation through  retrieval. Then, if retrieval is needed, the model generates a \textit{critique} token to evaluate the retrieved passage’s relevance. and a \textit{critique} token to evaluate if the information in the response is supported by the retrieved passage. Finally, a \textit{critique} token evaluates the overall utility of the response and selects the optimal result as the final output.
\subsection{Tuning Pattern}
RAG is continuously integrating with more LLM-related technologies. In Modular RAG, many components are composed of trainable language models. Through fine-tuning, the performance of the components and the compatibility with the overall flow can be further optimized. This section will introduce three main patterns of fine-tuning stages, namely retriever fine-tuning, generator fine-tuning, and dual fine-tuning.

\subsubsection{Retriever FT}
\begin{figure}
    \centering
    \includegraphics[width=\linewidth]{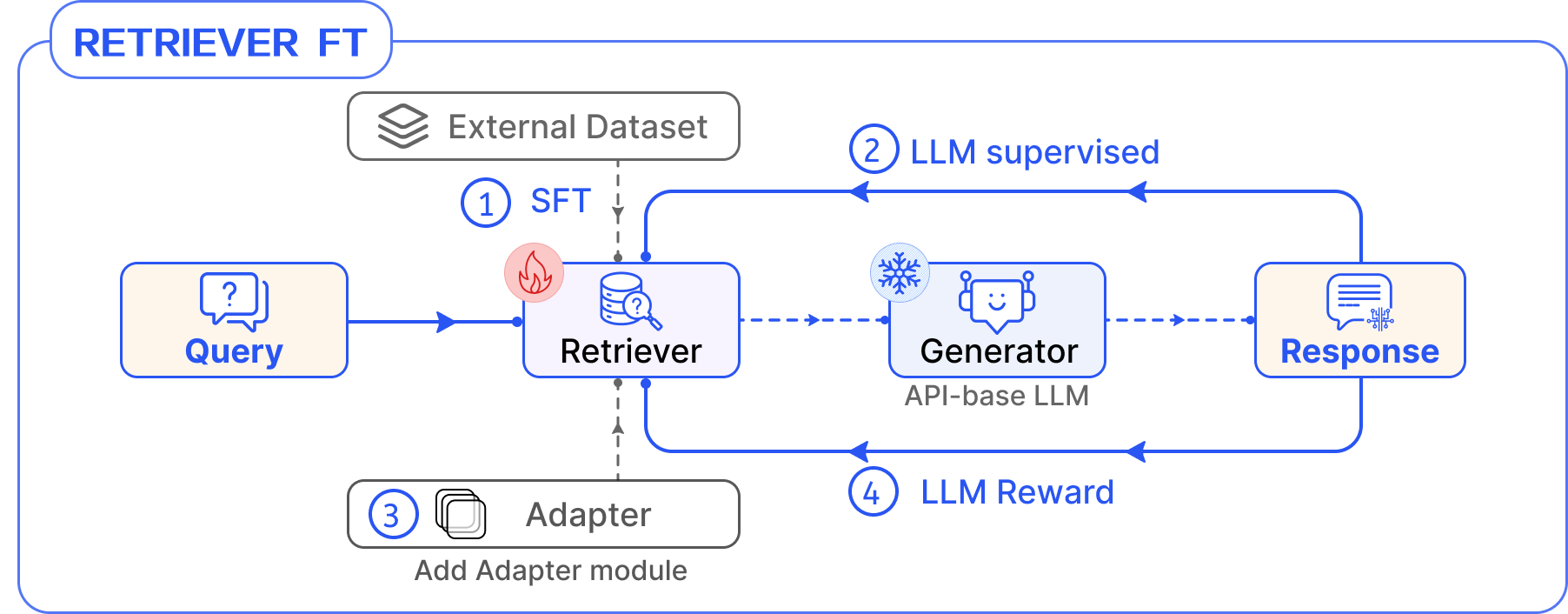}
    \caption{Retriever fine-tuning pattern, mainly includes direct SFT, adding trainable adapter, LM-supervised retrieval and LLM Reward RL.}
    \label{fig:Retriever_FT}
\end{figure}

In the RAG flow, common methods for fine-tuning the retriever is shown in Figure~\ref{fig:Retriever_FT} ,which include:
\begin{itemize}
    \item Direct supervised fine-tuning of the retriever. Constructing a specialized dataset for retrieval and fine-tuning the dense retriever. For example, using open-source retrieval datasets or constructing one based on domain-specific data.
    \item Adding trainable adapter modules. Sometimes, direct fine-tuning of the API-base embedding model (e.g., OpenAI Ada-002 and Cohere) is not feasible. Incorporating an adapter module can enhance the representation of your data. Additionally, the adapter module facilitates better alignment with downstream tasks, whether for task-specific (e.g., PRCA~\cite{PRCA}) or general purposes (e.g., AAR~\cite{AAR}).
    \item LM-supervised Retrieval (LSR). Fine-tuning the retriever based on the results generated by LLM.
    \item LLM Reward RL. Still using the LLM output results as the supervisory signal. Employing reinforcement learning to align the retriever with the generator. The whole retrieval process is disassembled in the form of a generative Markov chain.
\end{itemize}

\subsubsection{Generator FT}
\begin{figure}
    \centering
    \includegraphics[width=\linewidth]{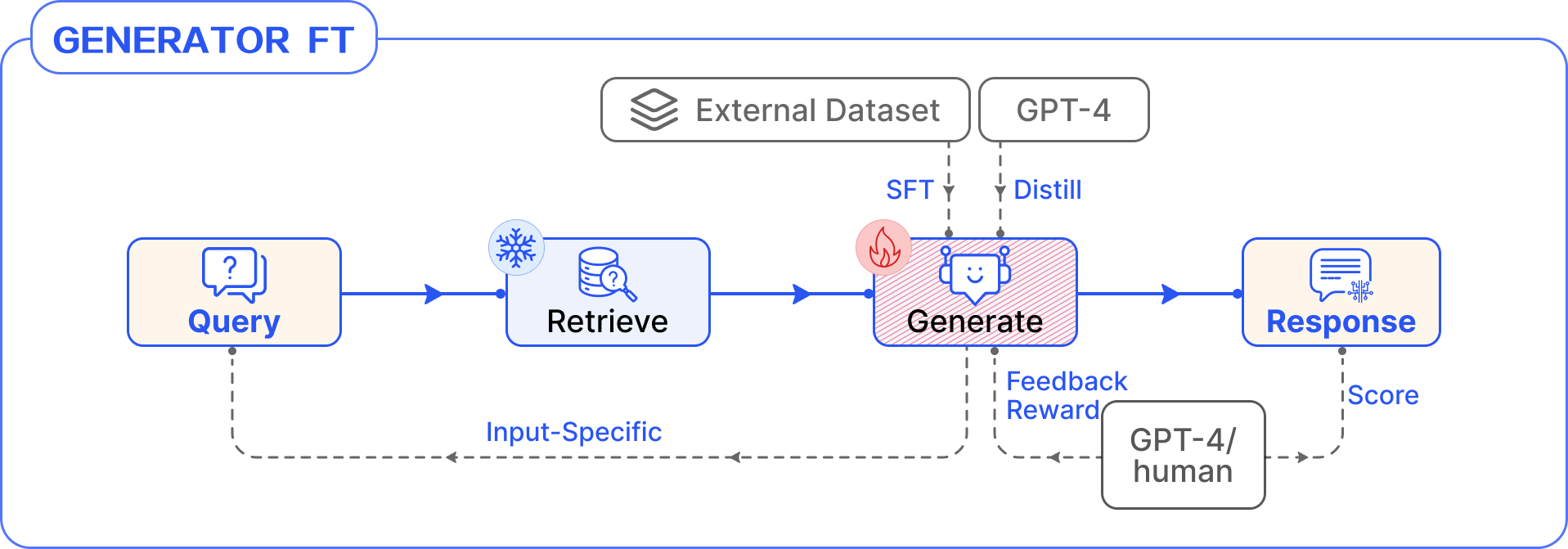}
    \caption{Generator fine-tuning pattern, The main methods include SFT, distillation and RL from LLM/human feedback.}
    \label{fig:Gen_FT}
\end{figure}

The primary methods for fine-tuning a generator in RAG flow is shown in Figure~\ref{fig:Gen_FT}, which include:
\begin{itemize}
    \item Direct supervised fine-tuning\textbf{.} Fine-tuning through an external dataset can supplement the generator with additional knowledge. Another benefit is the ability to customize input and output formats. By setting the Q\&A format, LLM can understand specific data formats and output according to instructions.
    \item Distillation\textbf{.} When using on-premise deployment of open-source models, a simple and effective Optimization method is to use GPT-4 to batch construct fine-tuning data to enhance the capabilities of the open-source model.
    \item RL from LLM/human feedback. Reinforcement learning based on feedback from the final generated answers. In addition to using human evaluations, powerful LLMs can also serve as an evaluative judge.
\end{itemize}

\subsubsection{Dual FT}
In the RAG system, fine-tuning both the retriever and the generator simultaneously is a unique feature of the RAG system. It is important to note that the emphasis of system fine-tuning is on the coordination between the retriever and the generator. An exemplary implementation is RA-DIT~\cite{RA-DIT}, which fine-tunes both the LLM and the retriever. The LM-ft component updates the LLM to maximize the likelihood of the correct answer given the retrieval-augmented instructions while the R-ft component updates the retriever to minimize the KL-Divergence between the retriever score distribution and the LLM preference.
 \begin{figure}
    \centering
    \includegraphics[width=\linewidth]{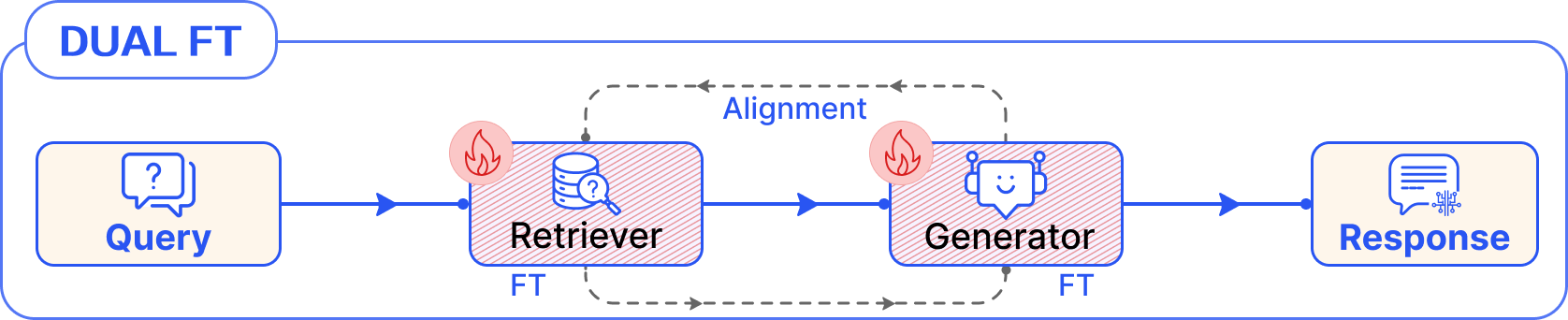}
    \caption{Dual fine-tuning pattern. In this mode, both the retriever and generator participate in fine-tuning, and their preferences will be aligned.}
    \label{fig:Dual_FT}
\end{figure}

\section{Discussion}
In this chapter, we explore the innovative horizons opened by the modular RAG paradigm. We examine its compatibility with cutting-edge methodologies in the progression of RAG technology, emphasizing its scalability. It not only fosters a fertile ground for model innovation but also paves the way for seamless adaptation to the dynamic requirements of various applications.

\subsection{Opportunities in Modular RAG}
The benefits of Modular RAG are evident, providing a fresh and comprehensive perspective on existing RAG-related work. Through modular organization, relevant technologies and methods are clearly summarized.

\textbf{From a research perspective.} Modular RAG is highly scalable, it empowers researchers to introduce innovative modules and operators, leveraging a deep understanding of RAG's evolving landscape. This flexibility enables the exploration of new theoretical and practical dimensions in the field.

\textbf{From an application perspective}. The modularity of RAG systems simplifies their design and implementation. Users can tailor RAG flows to fit their specific data, use cases, and downstream tasks, enhancing the adaptability of the system to diverse requirements. Developers can draw from existing flow architectures and innovate by defining new flows and patterns that are tailored to various application contexts and domains. This approach not only streamlines the development process but also enriches the functionality and versatility of RAG applications.

\subsection{Compatibility with new methods}

Modular RAG paradigm demonstrates exceptional compatibility with new developments. To gain a deeper understanding of this, we list three typical scalability cases, which clearly shows that Modular RAG paradigm provides robust support and flexibility for the innovation and development of RAG technology.

\subsubsection{Recombination of the current modules}
In this scenario, no new modules or operators are proposed; rather, specific problems are addressed through the combination of existing modules.DR-RAG~\cite{dr-RAG} employs a two-stage retrieval strategy and classifier selection mechanism, incorporating a branching retrieval structure. In the first stage, retrieving chunks relevant to the query. In the second stage, the query is combined individually with each chunk retrieved in the first stage, and a parallel secondary retrieval is conducted. The retrieved content is then input into a classifier to filter out the most relevant dynamic documents. This ensures that the retrieved documents are highly relevant to the query while reducing redundant information. DR-RAG improved retrieval method significantly enhances the accuracy and efficiency of answers, bolstering RAG's performance in multi-hop question-answering scenarios.

\subsubsection{New flow without adding new operators.}
This refers to redesigning the processes for retrieval and generation to address more complex scenarios without proposing new modules. The core idea of PlanRAG~\cite{planrag} lies in its introduction of a preliminary planning stage, a crucial step that occurs before retrieval and generation. Initially, the system employs a judge module to assess whether the current context necessitates the formulation of a new plan or adjustments to an existing one. When encountering a problem for the first time, the system initiates the planning process, while in subsequent interactions, it decides whether to execute re-planning based on previous plans and retrieved data.

Next, the system devises an execution plan tailored to the query, treating this process as a logical decomposition of complex queries. Specifically, PlanRAG uses a query expansion module to extend and refine the query. For each derived sub-query, the system conducts targeted  retrieval. Following retrieval, another judge module evaluates the current results to decide whether further retrieval is required or if it should return to the planning stage for re-planning. Through this strategy, PlanRAG is able to handle complex decision-making problems that require multi-step data analysis more efficiently. 

\subsubsection{New flow derived from  new operators.}
New operators often introduce novel flow design, exemplified by Multi-Head RAG~\cite{multi-head}. Existing RAG solutions do not focus on queries that may require retrieving multiple documents with significantly different content. Such queries are common but difficult to handle because embeddings of these documents may be far apart in the embedding space. Multi-Head RAG~\ addresses this by designing a new retriever that uses the activations of the multi-head attention layers of the Transformer, rather than the decoder layers, as keys for retrieving multifaceted documents. Different attention heads can learn to capture different aspects of the data. By using the corresponding activation results, embeddings that represent different aspects of the data items and the query can be generated, thereby enhancing the retrieval accuracy for complex queries.

\section{Conclusion}
RAG is emerging as a pivotal technology for LLM applications. As technological landscapes evolve and the intricacies of application requirements escalate, RAG systems are being enhanced by integrating a diverse suite of technologies, thereby achieving a higher level of complexity and functionality. This paper introduces the innovative paradigm of Modular RAG. This approach systematically disassembles the complex architecture of RAG systems into well-defined, discrete functional modules. Each module is meticulously characterized by its specific operational functions, ensuring clarity and precision. Therefore, the entire system is composed of those modules and operators, akin to Lego bricks. By conducting an in-depth analysis of numerous studies, the paper also distills common RAG design patterns and scrutinizes key case studies to illustrate these patterns in practice.

Modular RAG not only offers a structured framework for the design and application of RAG systems but also enables a scenario-based customization of these systems. The modularity inherent in this design facilitates ease of tracking and debugging, significantly enhancing the maintainability and scalability of RAG systems. Furthermore, Modular RAG opens up new avenues for the future progression of RAG technology. It encourages the innovation of novel functional modules and the crafting of innovative workflows, thereby driving forward the frontiers of RAG systems.

\bibliographystyle{IEEEtran}

\bibliography{modular}

\vfill

\end{document}